\documentclass[journal,twoside]{IEEEtran}

\usepackage{graphicx,amsmath,amsfonts,amssymb,bm,booktabs,threeparttable,subcaption}
\usepackage{hyperref,cleveref}
\hypersetup{colorlinks=true,linkcolor=[rgb]{0.1,0.3,0.9},urlcolor=[rgb]{0.2,0.2,0.2},citecolor=[rgb]{0.1,0.3,0.9}}
\DeclareMathAlphabet{\mathpzc}{OT1}{pzc}{m}{it}
\makeatletter\newcommand*\mysize{\@setfontsize\mysize{9.5}{11}}\makeatother
\def\rot#1{\rotatebox{90}{#1}}

\begin{document}

\title{CARRNN: A Continuous Autoregressive Recurrent Neural Network for Deep Representation Learning from Sporadic Temporal Data}

\author{Mostafa~Mehdipour~Ghazi, Lauge~S{\o}rensen, S\'ebastien~Ourselin, and~Mads~Nielsen
\thanks{\copyright 2021 IEEE. Personal use of this material is permitted. Permission from IEEE must be obtained for all other uses, in any current or future media, including reprinting/republishing this material for advertising or promotional purposes, creating new collective works, for resale or redistribution to servers or lists, or reuse of any copyrighted component of this work in other works.}
\thanks{The corresponding author M. Mehdipour Ghazi (ghazi@di.ku.dk) was with the Department of Medical Physics and Biomedical Engineering, University College London, London, UK, and Biomediq A/S, Copenhagen, DK. He is now with the Department of Computer Science, University of Copenhagen, Copenhagen, DK.}
\thanks{L. S{\o}rensen and M. Nielsen are with the Department of Computer Science, University of Copenhagen, Copenhagen, DK. They are also with Biomediq A/S and Cerebriu A/S, Copenhagen, DK.}
\thanks{S. Ourselin was with the Department of Medical Physics and Biomedical Engineering, University College London, London, UK. He is now with the School of Biomedical Engineering \& Imaging Sciences, King's College London, London, UK.}
\thanks{Data used in the preparation of this article were obtained from the Alzheimer's Disease Neuroimaging Initiative (ADNI) database (adni.loni.usc.edu). As such, the investigators within the ADNI contributed to the design and implementation of ADNI and/or provided data but did not participate in analysis or writing of this report. A complete listing of ADNI investigators can be found at \url{http://adni.loni.usc.edu/wp-content/uploads/how_to_apply/ADNI_Acknowledgement_List.pdf}}}


\IEEEtitleabstractindextext{
\begin{abstract}
Learning temporal patterns from multivariate longitudinal data is challenging especially in cases when data is sporadic, as often seen in, e.g., healthcare applications where the data can suffer from irregularity and asynchronicity as the time between consecutive data points can vary across features and samples, hindering the application of existing deep learning models that are constructed for complete, evenly spaced data with fixed sequence lengths. In this paper, a novel deep learning-based model is developed for modeling multiple temporal features in sporadic data using an integrated deep learning architecture based on a recurrent neural network (RNN) unit and a continuous-time autoregressive (CAR) model. The proposed model, called CARRNN, uses a generalized discrete-time autoregressive model that is trainable end-to-end using neural networks modulated by time lags to describe the changes caused by the irregularity and asynchronicity. It is applied to multivariate time-series regression tasks using data provided for Alzheimer's disease progression modeling and intensive care unit (ICU) mortality rate prediction, where the proposed model based on a gated recurrent unit (GRU) achieves the lowest prediction errors among the proposed RNN-based models and state-of-the-art methods using GRUs and long short-term memory (LSTM) networks in their architecture.
\end{abstract}

}

\maketitle
\IEEEdisplaynontitleabstractindextext
\IEEEpeerreviewmaketitle

\section{Introduction}

\IEEEPARstart{T}{he} rapid development of computational resources in recent years has enabled the processing of large-scale temporal sequences, especially in healthcare, using deep learning architectures. State-of-the-art multivariate sequence learning methods such as recurrent neural networks (RNNs) have been applied to extract high-level, time-dependent patterns from longitudinal data. Moreover, different variants of RNNs with gating mechanisms such as long short-term memory (LSTM) \cite{Hochreiter1997} and gated recurrent unit (GRU) \cite{Cho2014} were introduced to tackle the vanishing and exploding gradients problem \cite{Bengio1994,Hochreiter2001} and capture long-term dependencies efficiently.

\begin{figure*}[t]
\centering
\includegraphics[scale=0.7]{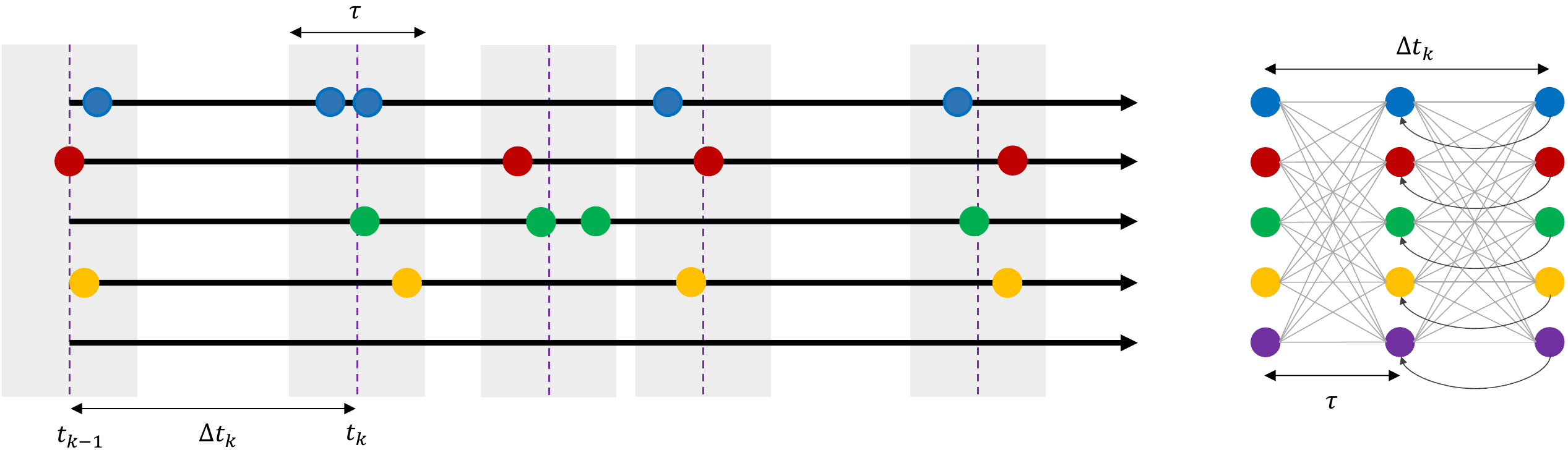}
\caption{Illustrative example of time-series data with irregularly-spaced time points and asynchronous events, and the proposed CARRNN structure for modeling the binned time points. The left subfigure shows five feature sequences of a sample subject aligned with a bin width of $\tau$ and a time gap of $\Delta t_{k}$ defined between the two consecutive data points at time $t_{k}$ and $t_{k-1}$. The right subfigure shows the CARRNN model structure that learns the multivariate temporal dependencies from the binned data. Note that the bottom feature includes only missing values.}
\label{fig_irregular}
\end{figure*}

However, RNNs are modeled as discrete-time dynamical systems with evenly-spaced input and output time points, thereby ill-suited to process sporadic data which is common in many healthcare applications \cite{Petersen2010,Silva2012}. This real-world data problem, as shown in the left part of Figure \ref{fig_irregular}, can, e.g., arise from different acquisition dates and missing values and is associated with irregularity and asynchronicity of the features where the time between consecutive timestamps or visits can vary across different features and subjects. To address this issue, most of the existing approaches \cite{Lipton2016} use a two-step process assuming a fixed interval and applying missing data imputation techniques to complete the data before modeling the time-series measurements using RNNs. On the other hand, a majority of methods that can inherently model varied-length data using RNNs without taking notice of the sampling time information \cite{Ghazi2019MedIA} fail to handle irregular or asynchronous data.

Recently, there have been efforts to incorporate the time distance attribute into the RNN architectures for dealing with sporadic data. For instance, the PLSTM \cite{Neil2016}, T-LSTM \cite{Baytas2017}, GRU-D \cite{Che2018}, tLSTM \cite{Santeramo2018}, and DLSTM \cite{Gao2019} have placed exponential or linear time gates in LSTM or GRU architectures heuristically as multiplicative time-modulating functions with learnable parameters and achieved good results when applied to sporadic data. However, none of these studies have analytically investigated the proposed models, nor have they provided motivation for the RNN architecture modifications. In other words, current studies mainly focus on the design of deep architectures or loss functions or apply the proposed solutions to only one specific type of RNN while the capability of the method using different types of RNNs is not examined.

\begin{figure}[t]
\centering
\includegraphics[scale=0.63]{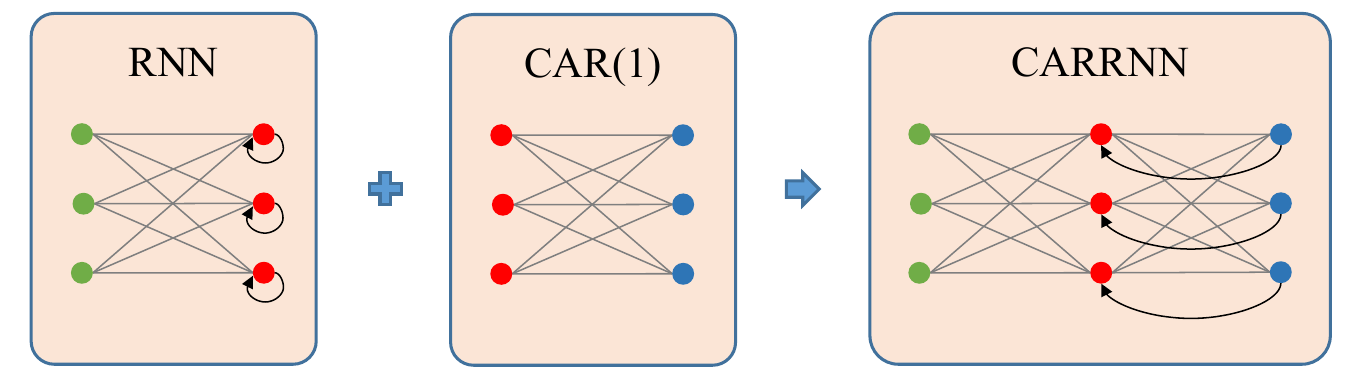}
\caption{Illustration of the proposed continuous-time autoregressive recurrent neural network model using an RNN and a CAR(1) model simulated as a built-in linear neural network layer for end-to-end learning.}
\label{fig_carrnn}
\end{figure}

In this paper, a novel model is reported for modeling multiple temporal features in sporadic multivariate longitudinal data using an integrated deep learning architecture as represented in Figure \ref{fig_carrnn} based on an RNN, LSTM, or GRU to learn the long-term dependencies from evenly-spaced data and a continuous-time autoregressive (CAR) model implemented as a neural network layer modulated by time lags to compensate for the existing irregularities. The proposed model, called CARRNN, introduces an analytical solution to the ordinary differential equation (ODE) problem and utilizes a generalized discrete-time autoregressive model which is trainable end-to-end. The asynchronous events are aligned using data binning to reduce the effects of missing values and noise. Finally, the remaining missing values are estimated by using a univariate CAR(1) model applied to the adjacent observations, and in cases when the feature points are fully missing, as shown in the bottom of the left part of Figure \ref{fig_irregular}, the input array and loss function are normalized with the number of available data points \cite{Ghazi2019MedIA}.

\section{The Proposed Method}

The proposed continuous-time autoregressive recurrent neural network model, which is trainable end-to-end as shown in Figure \ref{fig_carrnn}, applies an RNN to learn the long-term dependencies from evenly-spaced temporal data and uses a CAR(1) model as a neural network layer to compensate for the existing irregularities.

\begin{figure*}[t]
\centering
\includegraphics[scale=0.83]{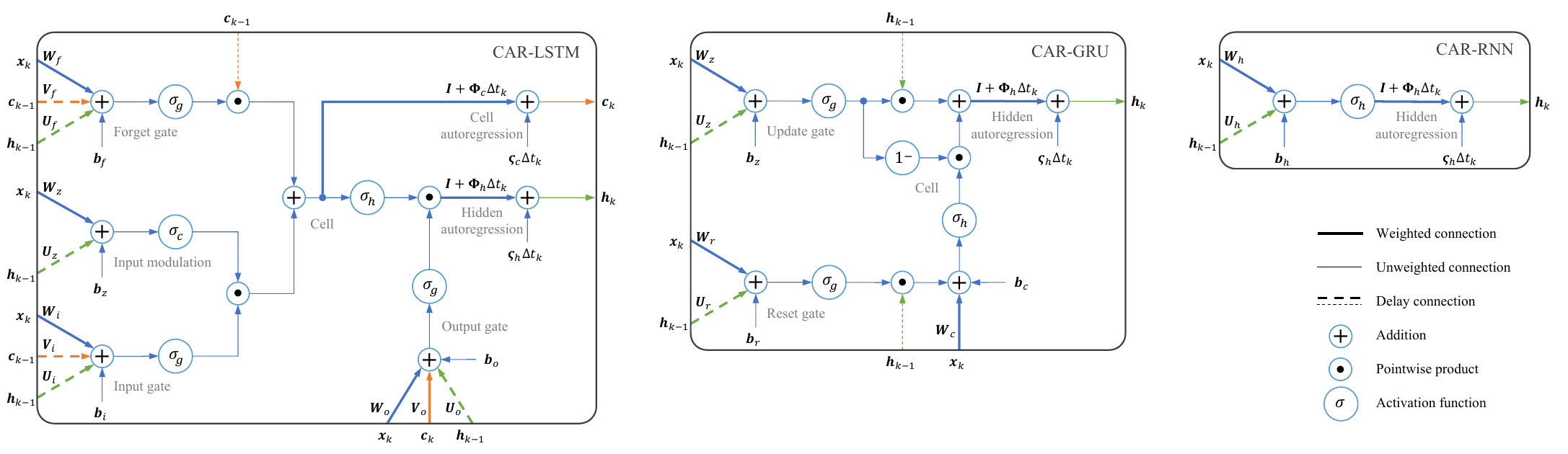}
\caption{Illustration of the proposed network architectures with built-in continuous autoregressive models.}
\label{fig_architecs}
\end{figure*}

\subsection{CAR Model}

Assume that $\bm{x}_{t} \in \mathbb{R}^{N \times 1}$ is a feature vector at time $t$ and can be predicted using a multivariate discrete autoregressive (AR) model of order one based on its evenly-spaced previous observations as follows
\begin{gather*}
\bm{x}_{t} = \bm{\phi} \bm{x}_{t-1} + \bm{c} + \bm{\varepsilon}_{t} \,,
\end{gather*}
\noindent where $\bm{\phi} \in \mathbb{R}^{N \times N}$ is the matrix of the slope coefficients of the model, $\bm{c} \in \mathbb{R}^{N \times 1}$ is the vector of the regression constants or intercepts, and $\bm{\varepsilon}_{t} \in \mathbb{R}^{N \times 1}$ is white noise as the prediction error. In general, the time interval between the consecutive observations can vary, making the sequence seem like an irregularly-sampled data. Therefore, the aforementioned AR(1) model can be rewritten as
\begin{gather*}
\bm{x}(t_{k}) = \bm{\phi}(\Delta t_{k}) \bm{x}(t_{k-1}) + \bm{c}(\Delta t_{k}) + \bm{\varepsilon}(t_{k}) \,,
\end{gather*}
\noindent where $\Delta t_{k} = t_{k} - t_{k-1}$ is the time gap between the two consecutive data points at time $t_{k}$ and $t_{k-1}$, as shown in Figure \ref{fig_irregular}, and the matrix $\bm{\phi}(\cdot)$, modulated by $\Delta t_{k}$, contains the autoregressive effects in the main diagonal and cross-lagged effects in the off-diagonals. The model can then be expressed by the first-order differential equation \cite{Voelkle2012,Haan2017} using continuous-time autoregressive parameters (drift matrix and bias vector) of $\bm{\Phi} \in \mathbb{R}^{N \times N}$ and $\bm{\varsigma} \in \mathbb{R}^{N \times 1}$ with an exponential solution as
\begin{gather*}
\frac{d \bm{x}(t)}{dt} = \bm{\Phi} \bm{x}(t) + \bm{\varsigma} + \bm{\Gamma} \frac{d \bm{\varepsilon}(t)}{dt} \,, \\
\bm{x}(t_{k}) = \mathrm{e}^{\bm{\Phi} \Delta t_{k}} \bm{x}(t_{k-1}) + \bm{\Phi}^{-1} \big[\mathrm{e}^{\bm{\Phi} \Delta t_{k}} - \bm{I}_N\big] \bm{\varsigma} + \bm{\eta}(t_{k}) \,,
\end{gather*}
\noindent where $\bm{I}_N$ is the identity matrix of size $N \times N$, $\bm{\Gamma} \in \mathbb{R}^{N \times N}$ is the Cholesky triangle of the innovation covariance or diffusion matrix $\bm{\Psi}$ ($\bm{\Psi} = \bm{\Gamma} \bm{\Gamma}^{\mathsf{T}}$, where ${\mathsf{T}}$ is the transpose operator), and $\bm{\eta}(t_{k}) \in \mathbb{R}^{N \times 1}$ is the continuous-time error vector at time $t_{k}$ which can be obtained as
\begin{gather*}
\bm{\eta}(t_{k}) = \int_{t_{k-1}}^{t_{k}} \mathrm{e}^{(t_{k} - s) \bm{\Phi}} \bm{\Gamma} d \bm{\varepsilon}(s) \,.
\end{gather*}

To avoid evaluating the matrix exponential function and its derivative, a power-series expansion can be used as follows
\begin{gather}
\nonumber \mathrm{e}^{\bm{\Phi} \Delta t_{k}} = \sum_{p=0}^{\infty} \frac{(\bm{\Phi} \Delta t_{k})^p}{p!} \approx \bm{I}_N + \bm{\Phi} \Delta t_{k} \,, \\
\bm{x}(t_{k}) \approx \big[\bm{I}_N + \bm{\Phi} \Delta t_{k}\big] \bm{x}(t_{k-1}) + \bm{\varsigma} \Delta t_{k} + \bm{\eta}(t_{k}) \,. \label{car_model}
\end{gather}

\subsection{RNN Model}

Simple recurrent networks \cite{Elman1990} are widely used for sequence prediction tasks by storing the previous values of the hidden units based on the following equation
\begin{gather}
\bm{h}_{t} = \sigma_{h}(\bm{W}_{h} \bm{x}_{t} + \bm{U}_{h} \bm{h}_{t-1} + \bm{b}_{h}) \,, \label{rnn_model} \\
\nonumber \bm{y}_{t} = \sigma_{y}(\bm{W}_{y} \bm{h}_{t} + \bm{b}_{y}) \,,
\end{gather}
\noindent where $\bm{h}_{t} \in \mathbb{R}^{M \times 1}$ and $\bm{y}_{t} \in \mathbb{R}^{Q \times 1}$ are the hidden (recurrent) and output layer vectors with $M$ and $Q$ nodes, respectively, at an evenly-spaced instant $t$, $\bm{W}_{h} \in \mathbb{R}^{M \times N}$ and $\bm{U}_{h} \in \mathbb{R}^{M \times M}$ are the input and hidden weight matrices with $N$ input nodes, $\bm{b}_{h} \in \mathbb{R}^{M \times 1}$ is the hidden bias vector, $\bm{W}_{y} \in \mathbb{R}^{Q \times M}$ and $\bm{b}_{y} \in \mathbb{R}^{Q \times 1}$ are the output weight matrix and bias vector, and $\sigma_{h}$ and $\sigma_{y}$ are the hidden and output layer activation functions, respectively. In a regression problem, $Q$ is set equal to $N$ and an identity function is applied to $\sigma_{y}$, while in a classification task, $Q$ is chosen equal to the number of classes and a Softmax function is used for $\sigma_{y}$.

\subsection{CARRNN Model}

The proposed CARRNN model can be obtained based on an integration of the CAR(1) model and one of the recursive neural network types. To avoid confusion, hereinafter the CARRNN model using an RNN, LSTM, and GRU are referred to as CAR-RNN, CAR-LSTM, and CAR-GRU, respectively.

\subsubsection{\textbf{CAR-RNN}}

Taking the advantages of the CAR(1) model introduced in Equation (\ref{car_model}) and the deep learning-based RNN model in Equation (\ref{rnn_model}) into account, the proposed learning-based model for irregularly-spaced sequence prediction can be obtained as follows
\begin{gather}
\nonumber \tilde{\bm{h}}_{k} = \sigma_{h}(\bm{W}_{h} \bm{x}_{k} + \bm{U}_{h} \bm{h}_{k-1} + \bm{b}_{h}) \,, \\
\bm{h}_{k} = \big[\bm{I}_M + (\Delta t_{k} - \tau) \bm{\Phi}_{h}\big] \tilde{\bm{h}}_{k} + (\Delta t_{k} - \tau) \bm{\varsigma}_{h} \,, \label{carrnn_model}
\end{gather}
\noindent where $\tilde{\bm{h}}_{k} \in \mathbb{R}^{M \times 1}$ is the regularized recurrent vector about time point $t_{k}$, $\bm{\Phi}_{h} \in \mathbb{R}^{M \times M}$ and $\bm{\varsigma}_{h} \in \mathbb{R}^{M \times 1}$ are the autoregressive weight matrix and bias vector, and $\Delta t_{\mathrm{min}}\leq \tau \leq \Delta t_{\mathrm{max}}$ is the RNN time-step hyperparameter. The right subfigure in figure \ref{fig_architecs} shows a schematic of the proposed CARRNN architecture.

It should be noted that when data is evenly-spaced, i.e., $\Delta t_{k} = \tau$, the model is simply generalized to a standard RNN. In addition, to train the model in an end-to-end fashion, as it can also be seen in Figure \ref{fig_carrnn}, CAR(1) model is simulated as a linear neural network layer with parameters modulated (multiplied) by the time lags. Also, although most of the current deep learning frameworks can approximate the parameter gradients using the automatic differentiation, we provide the details of calculating the gradients in Appendix \ref{append_gradients} for a fast and accurate evaluation of them during the network training.

\subsubsection{\textbf{CAR-LSTM}}

The abovementioned model can be generalized to the LSTM units for long-term prediction of unevenly-spaced sequences. To this end, the feedforward pass of the proposed model using a peephole LSTM unit \cite{Gers2002} can be expressed as
\begin{gather*}
\tilde{\bm{f}}_{k} = \sigma_{g}(\bm{W}_{f} \bm{x}_{k} + \bm{U}_{f} \bm{h}_{k-1} + \bm{V}_{f} \bm{c}_{k-1} + \bm{b}_{f}) \,, \\
\tilde{\bm{i}}_{k} = \sigma_{g}(\bm{W}_{i} \bm{x}_{k} + \bm{U}_{i} \bm{h}_{k-1} + \bm{V}_{i} \bm{c}_{k-1} + \bm{b}_{i}) \,, \\
\tilde{\bm{z}}_{k} = \sigma_{c}(\bm{W}_{z} \bm{x}_{k} + \bm{U}_{z} \bm{h}_{k-1} + \bm{b}_{z}) \,, \\
\tilde{\bm{c}}_{k} = \sigma_{h}(\tilde{\bm{f}}_{k} \odot \bm{c}_{k-1} + \tilde{\bm{i}}_{k} \odot \tilde{\bm{z}}_{k}) \,, \\
\tilde{\bm{o}}_{k} = \sigma_{g}(\bm{W}_{o} \bm{x}_{k} + \bm{U}_{o} \bm{h}_{k-1} + \bm{V}_{o} \bm{c}_{k} + \bm{b}_{o}) \,, \\
\tilde{\bm{h}}_{k} = \tilde{\bm{o}}_{k} \odot \tilde{\bm{c}}_{k} \,,
\end{gather*}
\noindent where the hidden vector $\bm{h}_{k}$ and cell state $\bm{c}_{k}$ can be obtained in a similar way as mentioned in Equation (\ref{carrnn_model}) using $\bm{\Phi}_{h}$, $\bm{\varsigma}_{h}$, $\bm{\Phi}_{c}$, and $\bm{\varsigma}_{c}$ as the parameters and $\tilde{\bm{h}}_{k}$ and $\tilde{\bar{\bm{c}}}_{k}$ as input vectors to the regularization functions in (\ref{carrnn_model}), where the latter stands for the regularized cell state before activation about time point $t_k$. As can be seen, no activation function is applied to the recurrent vector in the LSTM unit. Also, $\{\tilde{\bm{f}}_{k},\tilde{\bm{i}}_{k},\tilde{\bm{z}}_{k},\tilde{\bm{c}}_{k},\tilde{\bm{o}}_{k},\tilde{\bm{h}}_{k}\} \in \mathbb{R}^{M \times 1}$ are the regularized vectors of forget gate, input gate, modulation gate, cell state, output gate, and hidden layer about time point $t_{k}$, respectively. In addition, $\{\bm{W}_{f},\bm{W}_{i},\bm{W}_{z},\bm{W}_{o}\} \in \mathbb{R}^{M \times N}$ are weight matrices connecting the LSTM input to the gates, $\{\bm{U}_{f},\bm{U}_{i},\bm{U}_{z},\bm{U}_{o}\} \in \mathbb{R}^{M \times M}$ are weight matrices connecting the recurrent input to the gates, $\{\bm{V}_f,\bm{V}_i,\bm{V}_o\}\in\mathbb{R}^{M \times M}$ are diagonal weight matrices connecting the cell to the gates and can be set to zeros in case of using a standard LSTM unit, $\{\bm{b}_{f},\bm{b}_{i},\bm{b}_{z},\bm{b}_{o}\} \in \mathbb{R}^{M \times 1}$ denote corresponding biases of the nodes, and $\odot$ is the Hadamard product. The activation functions allocated to the gates, input modulation, and hidden layer are represented by $\sigma_{g}$, $\sigma_{c}$, and $\sigma_{h}$, respectively.

\subsubsection{\textbf{CAR-GRU}}

Likewise, the generalized model can be applied to the GRUs. The feedforward pass of the proposed model using a GRU \cite{Cho2014} can be expressed as
\begin{gather*}
\tilde{\bm{z}}_{k} = \sigma_{g}(\bm{W}_{z} \bm{x}_{k} + \bm{U}_{z} \bm{h}_{k-1} + \bm{b}_{z}) \,, \\
\tilde{\bm{r}}_{k} = \sigma_{g}(\bm{W}_{r} \bm{x}_{k} + \bm{U}_{r} \bm{h}_{k-1} + \bm{b}_{r}) \,, \\
\tilde{\bm{c}}_{k} = \sigma_{h}(\bm{W}_{c} \bm{x}_{k} + \bm{U}_{c} (\tilde{\bm{r}}_{k} \odot \bm{h}_{k-1}) + \bm{b}_{c}) \,, \\
\tilde{\bm{h}}_{k} = (1 - \tilde{\bm{z}}_{k}) \odot \tilde{\bm{c}}_{k} + \tilde{\bm{z}}_{k} \odot \bm{h}_{k-1} \,,
\end{gather*}
\noindent where hidden vector $\bm{h}_{k}$ can be obtained using Equation (\ref{carrnn_model}). Also, $\{\tilde{\bm{z}}_{k},\tilde{\bm{r}}_{k},\tilde{\bm{c}}_{k},\tilde{\bm{h}}_{k}\} \in \mathbb{R}^{M \times 1}$ are the regularized vectors of update gate, reset gate, candidate state, and hidden layer about time point $t_{k}$, respectively. In addition, $\{\bm{W}_{z},\bm{W}_{r},\bm{W}_{c}\} \in \mathbb{R}^{M \times N}$ are weight matrices connecting the GRU input to the gates and candidate state, $\{\bm{U}_{z},\bm{U}_{r},\bm{U}_{c}\} \in \mathbb{R}^{M \times M}$ are weight matrices connecting the recurrent input to the gates and candidate state, and $\{\bm{b}_{z},\bm{b}_{r},\bm{b}_{c}\} \in \mathbb{R}^{M \times 1}$ denote corresponding biases of the nodes.

\subsection{Time Binning}

Time binning is used to discretize and align the continuous features within small intervals (bins) \cite{Anumula2018}. This will reduce the effects of noise and missing values for effective learning of the multivariate temporal dependencies from asynchronous data using the abovementioned deep learning models. Therefore, as also illustrated in Figure \ref{fig_irregular}, the features are allowed to be matched within a given bin width which is equal to the RNN time step $\tau$.

\subsection{Handling Missing Values} \label{sec_fill}

The proposed CARRNN model can also be developed for estimating missing values of incomplete features during training. To be more precise, a diagonal autoregressive matrix can be used in Equation (\ref{car_model}) to impute the missing values of each feature based on its adjacent or previous observations. This can be interpreted as variants of the nearest-neighbor (NN) imputation \cite{Beretta2016} or the last-observation-carried-forward (LOCF) method \cite{Molnar2008}, but the time intervals of the consecutive points are taken into account to adjust the replicated values using an independent (univariate) version of Equation (\ref{car_model}) during training, i.e.,
\begin{gather*}
x_{n,k} = \big[1 + (t_{k} - t_{j}) \bm{\varphi}(n)\big] x_{n,j} + (t_{k} - t_{j}) \bm{\zeta}(n) \,,
\end{gather*}
\noindent where $\bm{\varphi} \in \mathbb{R}^{N \times 1}$ and $\bm{\zeta} \in \mathbb{R}^{N \times 1}$ are the univariate continuous autoregressive model parameters, and $x_{n,k}$ is the value of the $n$th feature at timestamp $t_{k}$ estimated based on its neighboring observation at timestamp $t_{j}$.

\begin{figure}[t]
\centering
\includegraphics[scale=0.8]{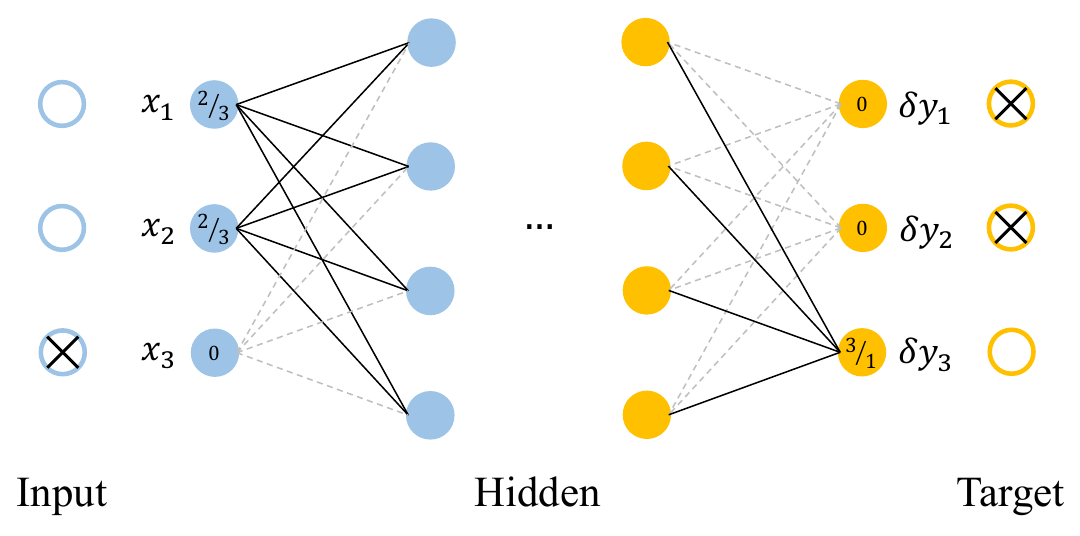}
\caption{Handling missing values using a weighted feedforward and backpropagation. In this example, $Q = N = 3$ and one input value and two output values are missing, indicated by crosses. The network nodes, and hence, their connected weights associated with the missing input and target values are set to zero, indicated by dashed lines, while the input values are scaled by $2 / 3$, which forms the ratio of the number of available input points to the total number of input features $N$, and the output gradients are scaled by $3 / 1$, which accounts for the ratio of the total number of target features $Q$ to the number of available target points.}
\label{fig_missing}
\end{figure}

However, missing values remains a problem in features with missing values at the beginning of the sequence or fully missing features in the input and target vectors. To deal with the remaining missing data, we use a weighted input array and loss function to regularize the network according to the number of available data points \cite{Ghazi2019MedIA}. This can be seen as the dropout technique \cite{Srivastava2014} where the network nodes are randomly skipped during training so that the network only learns and updates some of the weights per iteration. However, instead of a random selection of nodes, it is assumed that the missing nodes and their connected weights are dropped out from the learning process on purpose. This is equivalent to setting the input nodes associated with the missing input points to zero and multiplying the rest of the input values by the ratio of the number of available points in the input array per timestamp to the total number of input features $N$ during the feedforward process, and setting the output nodes associated with the missing target points to zero and multiplying the rest of the output gradients by the ratio of the total number of target features $Q$ to the number of available points in the target array per timestamp during the backpropagation procedure. Figure \ref{fig_missing} represents how the network nodes, and consequently, their connected weights are scaled to forward the input array and to propagate the output gradients associated with the available input and target values.

\section{Experiments and Results}

\begin{table*}[t]
\mysize
\centering
\renewcommand{\arraystretch}{1.3}
\caption{Statistics of the utilized datasets after cleaning.}
\begin{threeparttable}
\begin{tabular}{lccccc}
\toprule
 & \# subjects & Visit interval & \# features per visit & \# visits per subject \\
 & (stable | converting) & (mean$\pm$SD) | [min max] & (mean$\pm$SD) | [min max] & (mean$\pm$SD) | [min max] \\
\midrule
ADNI & 471 | 313 & 0.74$\pm$0.43 | [0.05 4.82] year & 9.78$\pm$3.38 | [1 16] & 5.99$\pm$2.37 | [2 13] \\
PhysioNet & 10,275 | 1,706 & 0.64$\pm$0.52 | [0.017 31] hour & 5.36$\pm$2.44 | [1 27] & 73.63$\pm$22.55 | [2 210] \\
\bottomrule
\end{tabular}
\begin{tablenotes}
\item {\footnotesize In ADNI, stable or converting refer to the patients with a baseline mild cognitive impairment diagnosis staying the same or converting to Alzheimer’s dementia in the later follow-ups, respectively. In PhysioNet, stable and converting denote the surviving and dead cases after at least two days from their admission to ICU.}
\end{tablenotes}
\end{threeparttable}
\label{data_stat}
\end{table*}

\subsection{Data}

Two real-world datasets are used to train different time-series models with irregularity and asynchronicity. These longitudinal datasets are multivariate and contain missing data. The first dataset is obtained from the Alzheimer’s Disease Neuroimaging Initiative (\href{http://adni.loni.usc.edu/data-samples/access-data/}{ADNI}) cohort \cite{Petersen2010} for disease progression modeling using multimodal biomarkers obtained from T1-weighted brain magnetic resonance imaging (MRI) and positron emission tomography (PET) scans, cerebrospinal fluid (CSF) data, and cognitive tests. The ADNI was launched in 2003 as a public-private partnership, led by principal investigator Michael W. Weiner, MD. The primary goal of ADNI has been to test whether serial MRI, PET, other biological markers, and clinical and neuropsychological assessment can be combined to measure the progression of mild cognitive impairment and early Alzheimer's disease. The data was preprocessed and cleaned based on the criteria presented in \cite{Ghazi2021}. The utilized data includes 16 temporal biomarkers acquired from 1,518 subjects (854 males and 664 females aged between 55 and 98) in 9,098 timestamps or visits between 2005 and 2017.

The second dataset is obtained from the PhysioNet Computing in Cardiology (\href{https://www.physionet.org/content/challenge-2012/1.0.0/}{PhysioNet/CinC}) challenge \cite{Silva2012,Goldberger2000} to predict mortality rates of in-hospital patients using their physiological measurements including laboratory (blood) results and vital signs recorded from the first 48 hours of intensive care unit (ICU) stays. We discarded three biomarkers (ventilation, cholesterol, and troponin-I) with constant values and a very limited number of data points, and removed subjects with less than two distinct timestamps for sequence learning purposes. The remaining 33 time-series variables are collected from 11,981 subjects (6,713 males and 5,257 females aged between 15 and 90) in 882,207 timestamps during the first two days of ICU admissions. Table \ref{data_stat} summarizes statistics of the used datasets after cleaning. Note that both datasets also include missing values after cleaning.

To facilitate future research in time-series modeling and comparison with the current study, all source code and data splits are available online at \href{https://github.com/Mostafa-Ghazi/CARRNN}{https://github.com/Mostafa-Ghazi/CARRNN}.

\subsection{Experimental Setup}

The proposed CARRNN models were applied to regression problems using an identity function, hyperbolic tangent, and logistic sigmoid as activation functions for $\sigma_h$, $\sigma_c$, and $\sigma_g$, respectively. Since initialization is a key for faster convergence and stability of deep network training, the network biases and autoregressive weights were initialized to zero, and values of the RNN weight matrices were selected according to the rules and assumptions proposed in \cite{Ghazi2019ICONIP}.

The data was standardized to have zero mean and unit variance per feature dimension, and time intervals were normalized with the interquartile range (IQR) of the timestamps. In ADNI, $80\%$ of the samples were randomly selected for training and validation, and the remaining $20\%$ were used for testing the unseen test subset. The mini-batch size was set to $90\%$ of the training samples, and the number of hidden nodes was set to 10 times the number of input nodes. In PhysioNet, 7,986 samples were allotted for training and validation, and 3,995 samples were assigned for testing the unseen test subset. The mini-batch size was set to $25\%$ of the training samples, and the number of hidden nodes was set to 5 times the number of input nodes.

The first to penultimate time points were utilized to estimate the second to last time points with a prediction horizon of one step using the following methods
\begin{itemize}
\item \textit{GRU-Mean}, a standard GRU with missing values filled in using the mean values \cite{Garcia2010}.
\item \textit{GRU-Forward}, a standard GRU with missing values filled in using the previous observations \cite{ Molnar2008}.
\item \textit{GRU-Concat}, a standard GRU with missing values filled in using the nearest neighbors \cite{Beretta2016}, and input features concatenated with corresponding time intervals.
\item \textit{GRU-D}, a state-of-the-art method \cite{Che2018} that uses a modified GRU to impute missing value by the weighted combination of the last observation, mean value, and recurrent component.
\item \textit{CAR-RNN}, the proposed CARRNN model using an RNN unit in the architecture.
\item \textit{CAR-LSTM}, the proposed CARRNN model using an LSTM unit in the architecture.
\item \textit{CAR-GRU}, the proposed CARRNN model using a GRU in the architecture.
\end{itemize}

A normalized L2-norm loss was used as the cost function and the Adam optimizer \cite{Kingma2014} was applied as a gradient descent-based optimization algorithm to update the network parameters with a gradient decay factor of $0.85$, a squared gradient decay factor of $0.95$, and a base learning rate of $5 \times 10^{-3}$. An L2-norm regularization was applied to the weights with a weight decay factor of $5 \times 10^{-5}$. Different values of the time step $\tau$ (histogram bin width) were examined, and a grid search was used to find the optimal values of the hyperparameters according to the validation set error across different experiments and methods. Hence, the reported results are based on the selected (optimal) parameter values subsequently applied to the unseen test subsets.

The networks were trained for at most 100 epochs in a 10-fold nested cross-validation setup using the early-stopping method \cite{Prechelt1998} with 10-iterations patience. Since both datasets contain outliers, beside  the mean squared error (MSE), the mean absolute error (MAE) was used \cite{Chai2014} to evaluate the modeling performance in terms of the absolute differences between actual and estimated values.

\begin{figure}[t]
\centering
\begin{subfigure}[t]{0.495\textwidth}
\raisebox{-\height}{\includegraphics[scale=0.6]{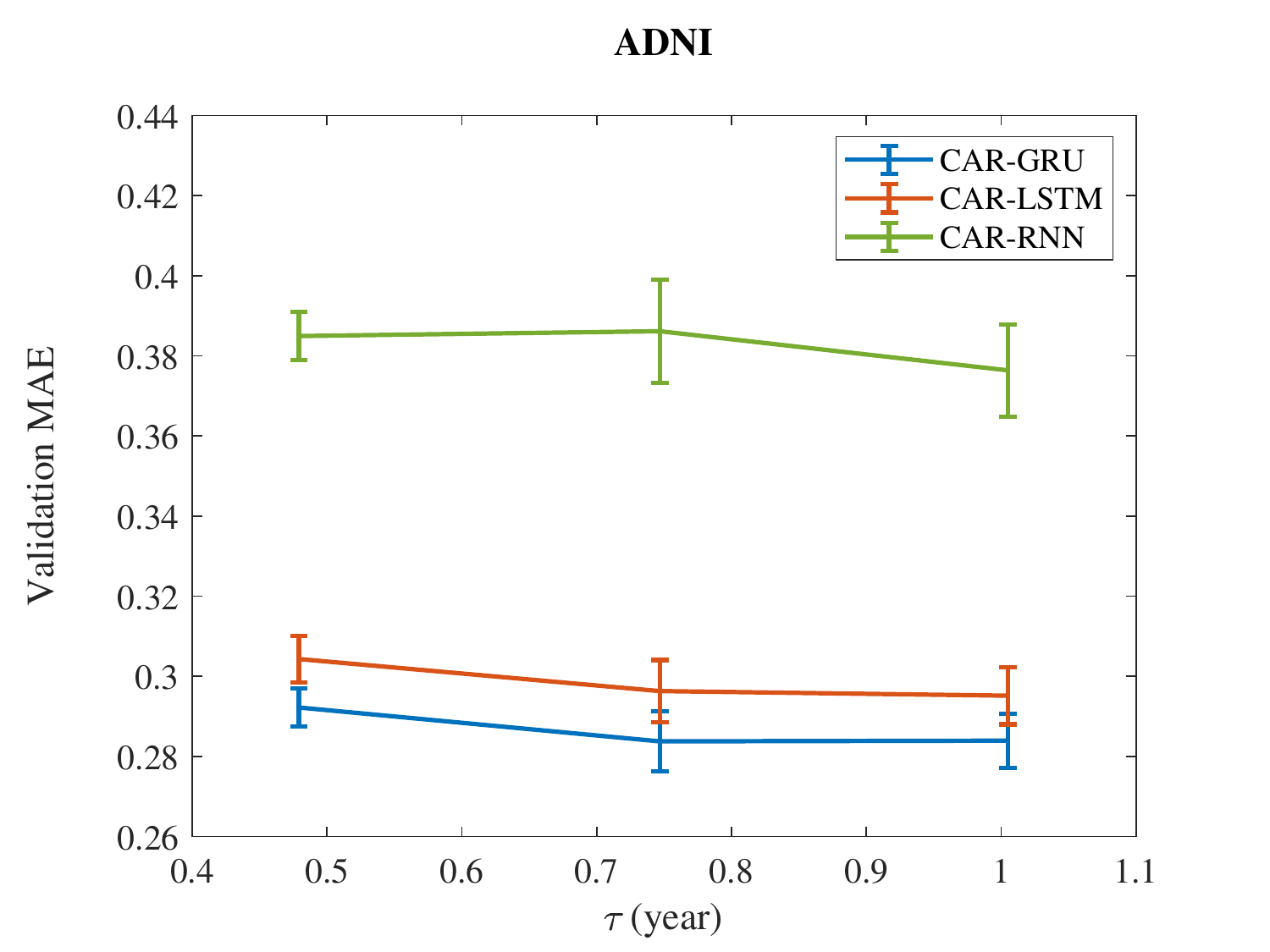}}
\end{subfigure}
\begin{subfigure}[t]{0.495\textwidth}
\raisebox{-\height}{\includegraphics[scale=0.6]{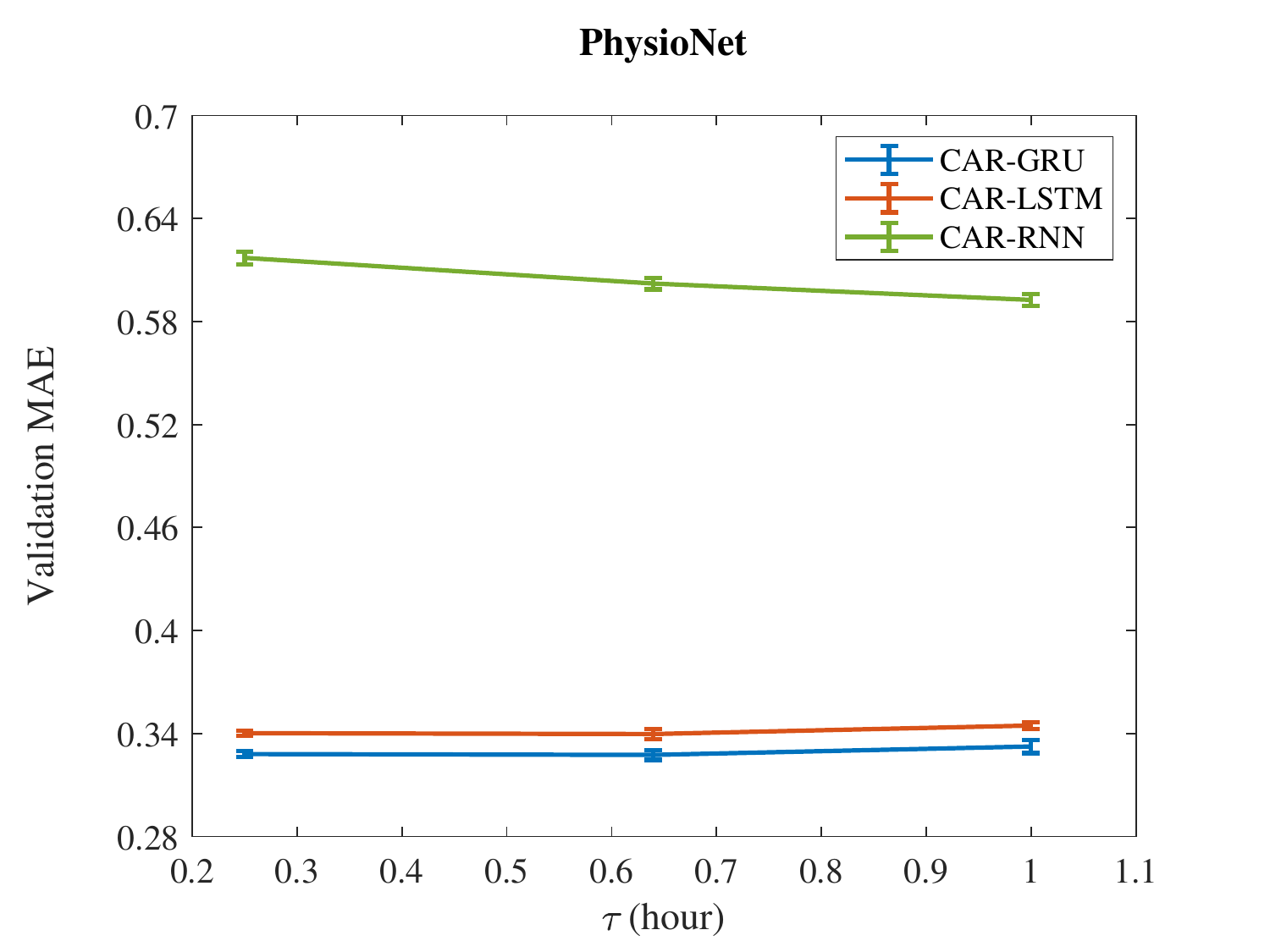}}
\end{subfigure}
\caption{Validation prediction performance for the proposed models applied to the utilized datasets with different time steps. The error bars are calculated based on a 95\% confidence interval for population standard deviation in 10-fold nested cross-validation per time step.}
\label{fig_valid_tau}
\end{figure}

\subsection{Results and Discussion}

\subsubsection{Validation performance for various time steps}

As the first set of experiments, we compare the validation prediction performance of the proposed models applied to the obtained datasets in a 10-fold nested cross-validation setup using different time steps ($\tau$) including the mean and IQR of the time intervals. Figure \ref{fig_valid_tau} shows the validation results of the proposed models for different time steps and datasets. As can be seen, all models show good stability to the variations of time steps when applied to both ADNI and PhysioNet. In all cases, CAR-GRU achieves the lowest prediction error, while CAR-RNN obtains the largest error with larger deviations across different runs, especially in PhysioNet which involves very long sequences. Moreover, except for the CAR-RNN, the optimal errors are achieved at the middle point where the time step is set to the average value of time intervals. These optimal values are selected to be used in the later experiments.

\subsubsection{Validation performance versus iteration}

\begin{figure}[t]
\centering
\begin{subfigure}[t]{0.495\textwidth}
\raisebox{-\height}{\includegraphics[scale=0.6]{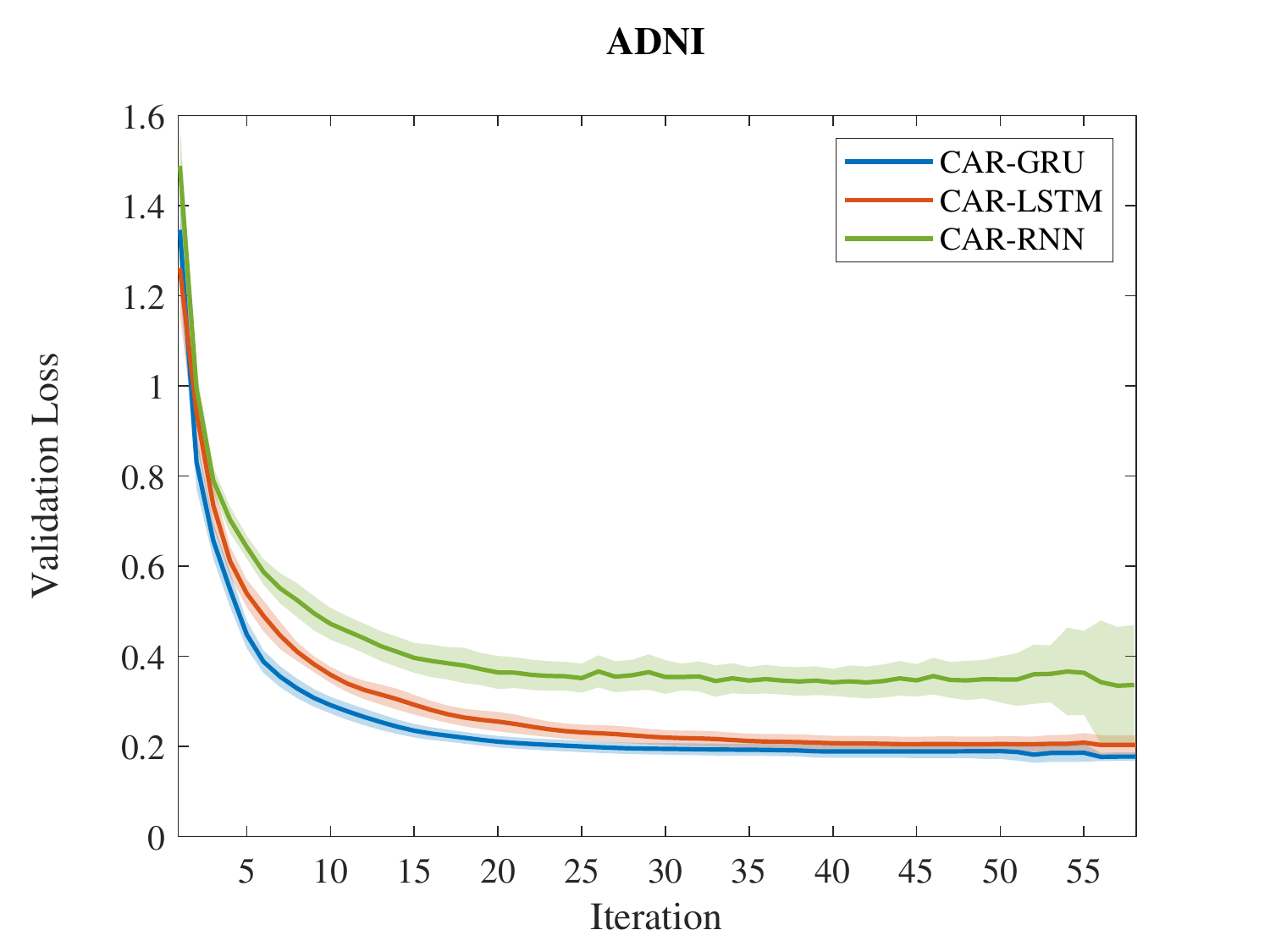}}
\end{subfigure}
\begin{subfigure}[t]{0.495\textwidth}
\raisebox{-\height}{\includegraphics[scale=0.6]{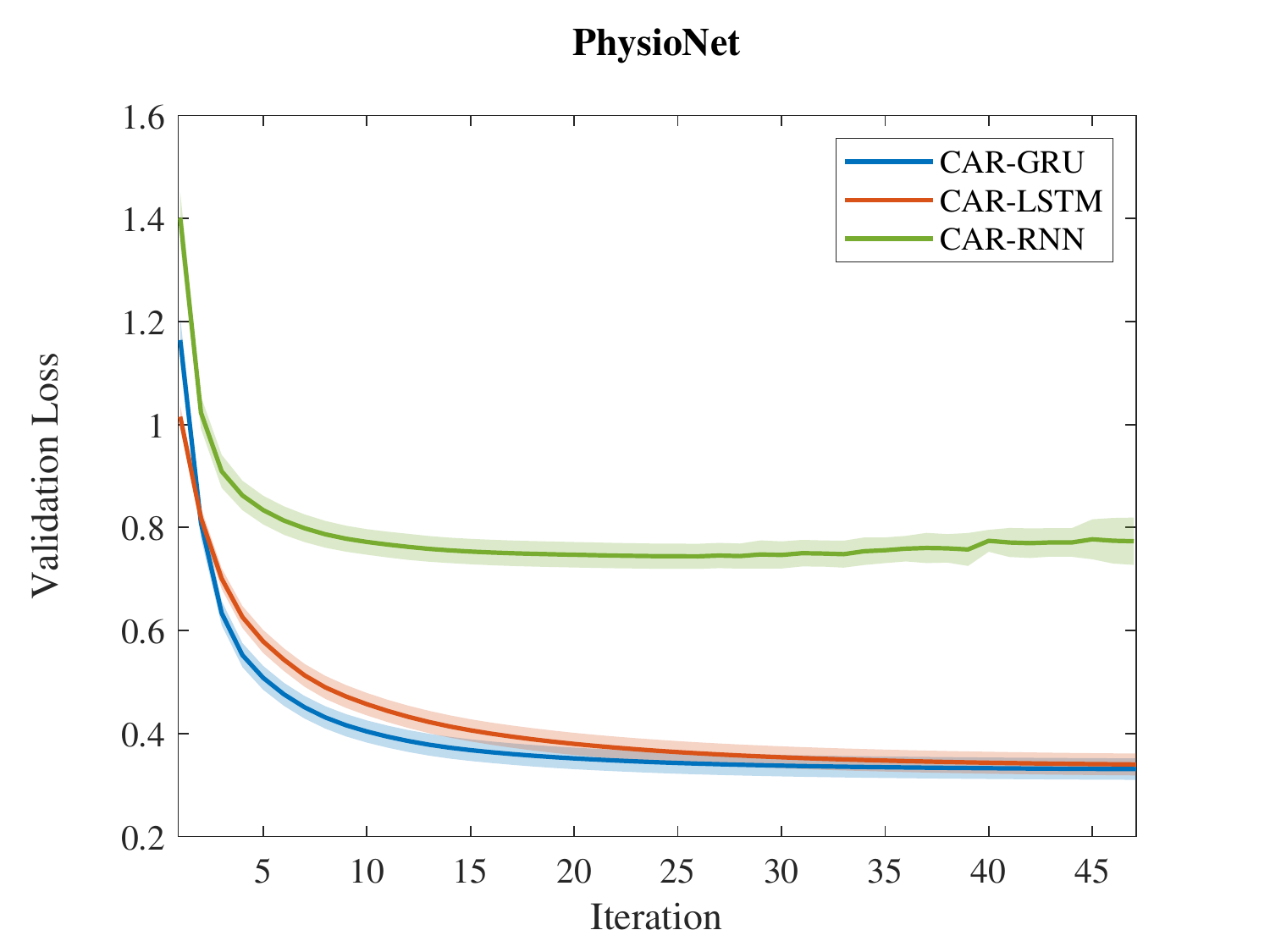}}
\end{subfigure}
\caption{Validation performance of the proposed models applied to the utilized datasets in the optimal time steps. The shaded areas display the 95\% confidence interval for population standard deviation of 10-fold nested cross-validation per iteration about the average curves.}
\label{fig_valid_iter}
\end{figure}

Additionally, we draw the validation loss of the proposed models monitored during the 10-times training with the early-stopping approach. Figure \ref{fig_valid_iter} demonstrates the tracked validation loss of the proposed models applied to the two datasets using the optimal time steps. It can be seen that the models are robust to initialization and data subsets with the lowest error achieved for the CAR-GRU. It can also be deduced that the CAR-RNN model cannot perform well for learning long-term dependencies, most probably due to the simplicity of its architecture and lack of gating structures to avoid exploding and vanishing gradients during backpropagation.

\begin{figure*}[t]
\centering
\begin{subfigure}[t]{0.505\textwidth}
\raisebox{-\height}{\includegraphics[scale=0.6]{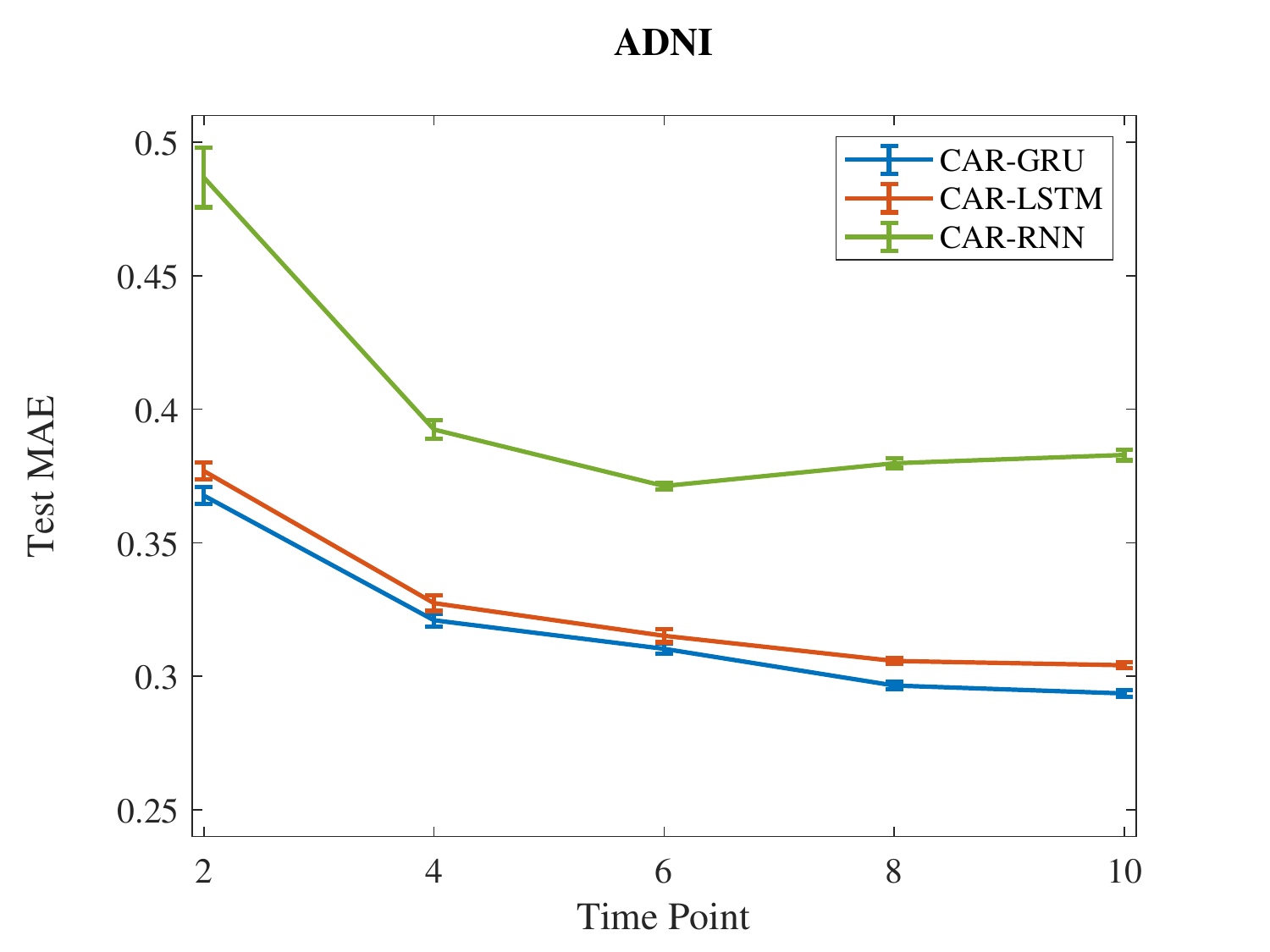}}
\end{subfigure}
\begin{subfigure}[t]{0.485\textwidth}
\raisebox{-\height}{\includegraphics[scale=0.6]{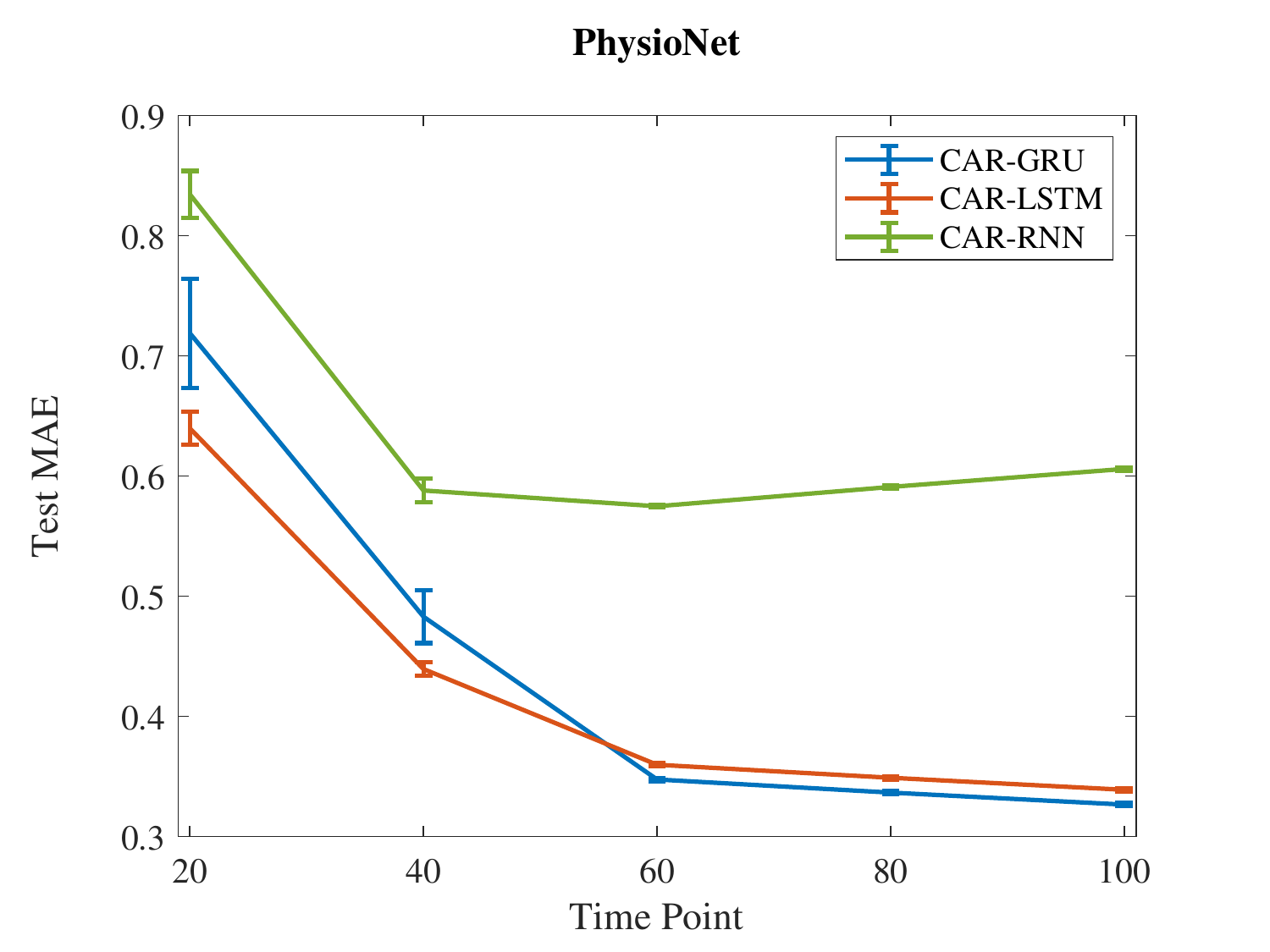}}
\end{subfigure}
\caption{Generalization performance of the proposed models applied to the utilized datasets in the optimal time steps based on different numbers of time points used for the follow-up predictions per subject. The error bars are calculated based on a 95\% confidence interval for population standard deviation in 10-fold nested cross-validation at each point.}
\label{fig_test_visit}
\end{figure*}

\begin{table*}[t]
\caption{Generalization errors (mean$\pm$SD) in predicting the test feature values using the trained models based on 10-fold nested cross-validation. The best results which are boldfaced have a statistically significant difference ($p < 0.05$) from those of the other methods.}
\label{table_results}
\mysize
\centering
\renewcommand{\arraystretch}{1.3}
\begin{tabular}{llccccccc}
\toprule
\multicolumn{2}{c}{Method} & GRU-Mean & GRU-Forward & GRU-Concat & GRU-D & CAR-RNN & CAR-LSTM & CAR-GRU \\
\midrule
 & ADNI & 0.453$\pm$0.007 & 0.323$\pm$0.002 & 0.304$\pm$0.002 & 0.315$\pm$0.002 & 0.386$\pm$0.004 & 0.297$\pm$0.002 & \textbf{0.286}$\pm$\textbf{0.002} \\
\rot{\rlap{MAE}} & PhysioNet & 0.546$\pm$0.006 & 0.456$\pm$0.003 & 0.334$\pm$0.002 & 0.371$\pm$0.002 & 0.606$\pm$0.002 & 0.339$\pm$0.002 & \textbf{0.326}$\pm$\textbf{0.002} \\
\midrule
 & ADNI & 0.373$\pm$0.011 & 0.206$\pm$0.002 & 0.189$\pm$0.002 & 0.181$\pm$0.002 & 0.319$\pm$0.006 & 0.178$\pm$0.003 & \textbf{0.167}$\pm$\textbf{0.002} \\
\rot{\rlap{MSE}} & PhysioNet & 0.589$\pm$0.008 & 0.494$\pm$0.003 & 0.375$\pm$0.002 & 0.402$\pm$0.002 & 0.739$\pm$0.003 & 0.319$\pm$0.002 & \textbf{0.308}$\pm$\textbf{0.002} \\
\bottomrule
\end{tabular}
\end{table*}

\subsubsection{Test performance for various time points}

Practically, it is important to see how the trained models would generalize to test data with various numbers of time points or visits. Therefore, we apply the cross-validated models to the test subsets using only a few visits of each subject to sequentially predict the later follow-ups. The results of this experiment using the optimal time steps are presented in Figure \ref{fig_test_visit}. As depicted in the figure, the trained models are generalizable to the test data to a very good extent, even when using very few time points per subject. Once again the CAR-GRU models obtain the lowest prediction errors among the other models in all cases.

\subsubsection{Comparison to the state-of-the-art}

Table \ref{table_results} compares the test results in predicting the feature values using different models applied to the two datasets. As can be seen, the CAR-GRU model achieves the lowest errors in modeling the sporadic data, and these results have a statistically significant difference ($p < 0.05$) from the findings of the other methods according to the two-sided Wilcoxon signed-rank sum test \cite{Wilcoxon1945}. Note that the methods are applied to the data after aligning the data points using the optimal time steps.

\subsubsection{Trajectory prediction and classification}

\begin{figure*}[t]
\centering
\begin{subfigure}[t]{0.505\textwidth}
\raisebox{-\height}{\includegraphics[scale=0.6]{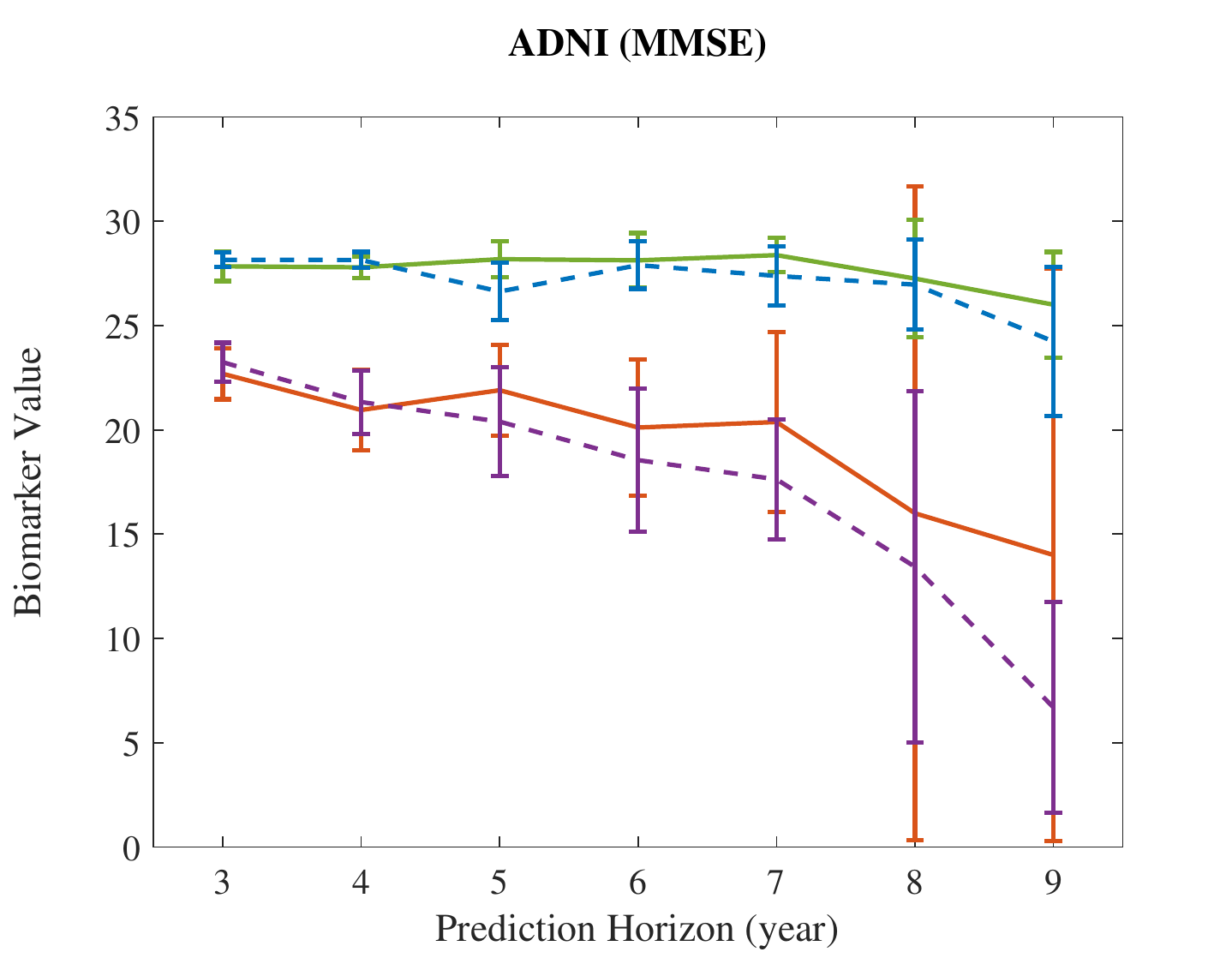}}
\end{subfigure}
\begin{subfigure}[t]{0.485\textwidth}
\raisebox{-\height}{\includegraphics[scale=0.6]{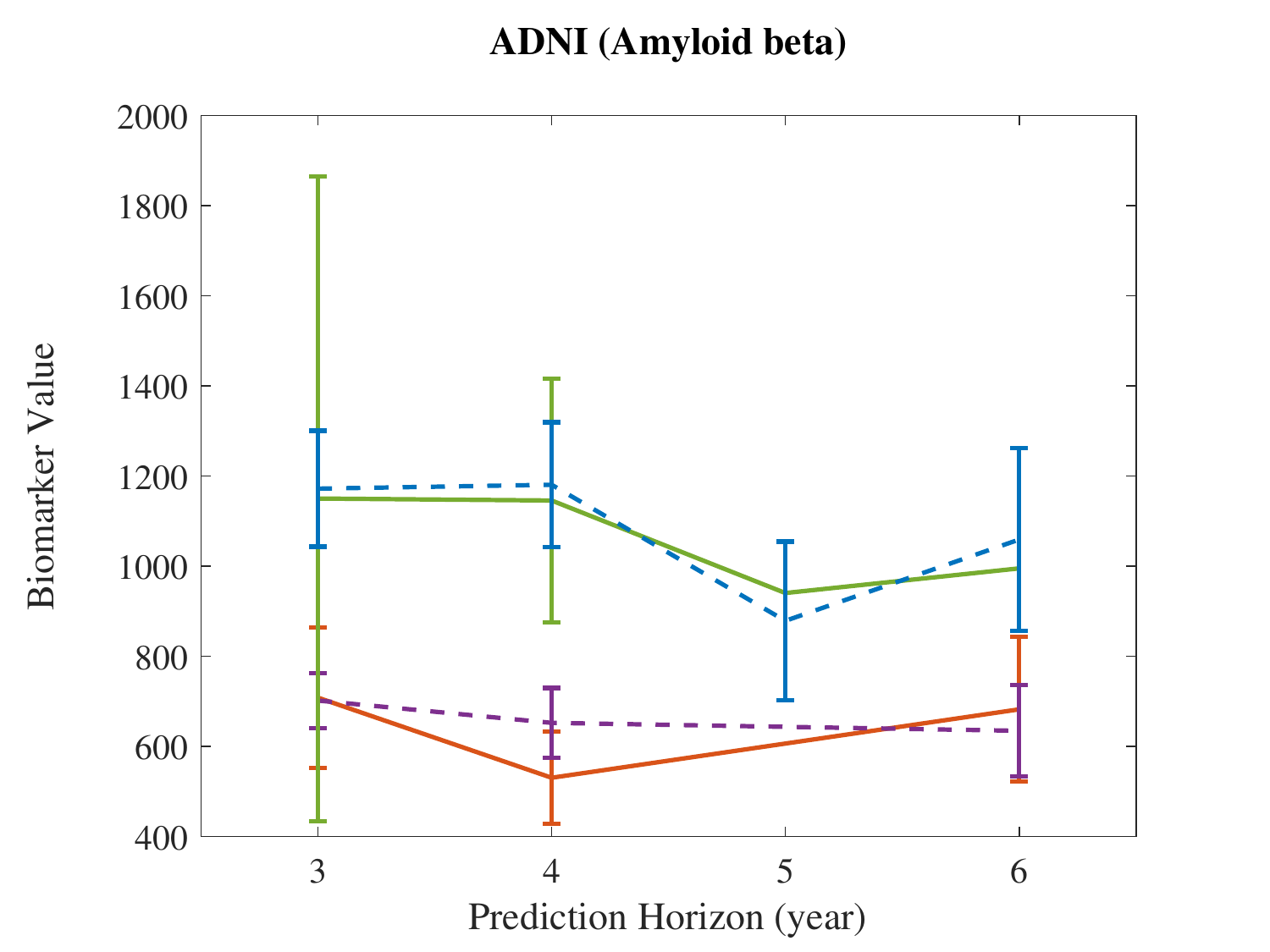}}
\end{subfigure}
\begin{subfigure}[t]{0.505\textwidth}
\raisebox{-\height}{\includegraphics[scale=0.6]{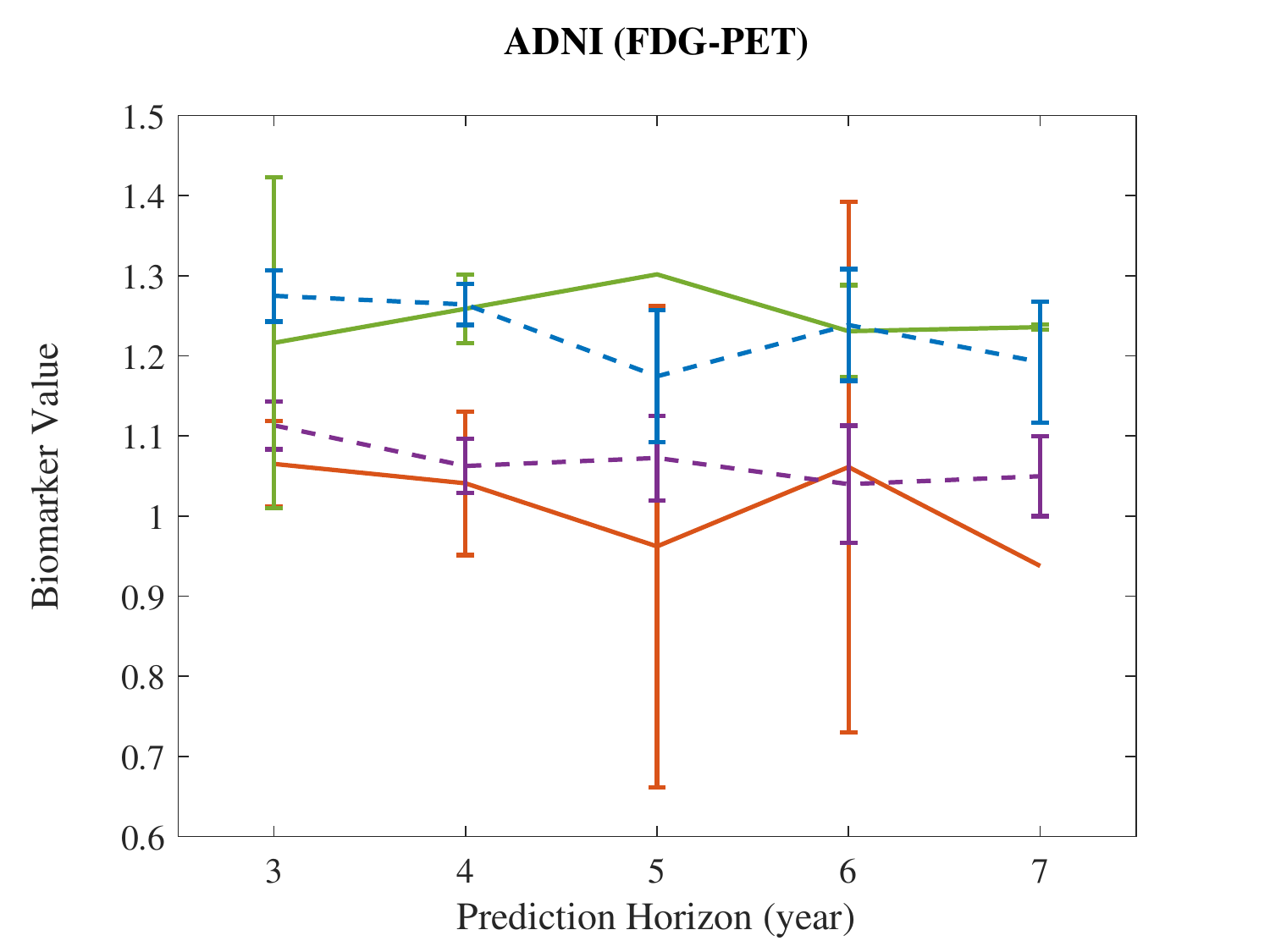}}
\end{subfigure}
\begin{subfigure}[t]{0.485\textwidth}
\raisebox{-\height}{\includegraphics[scale=0.6]{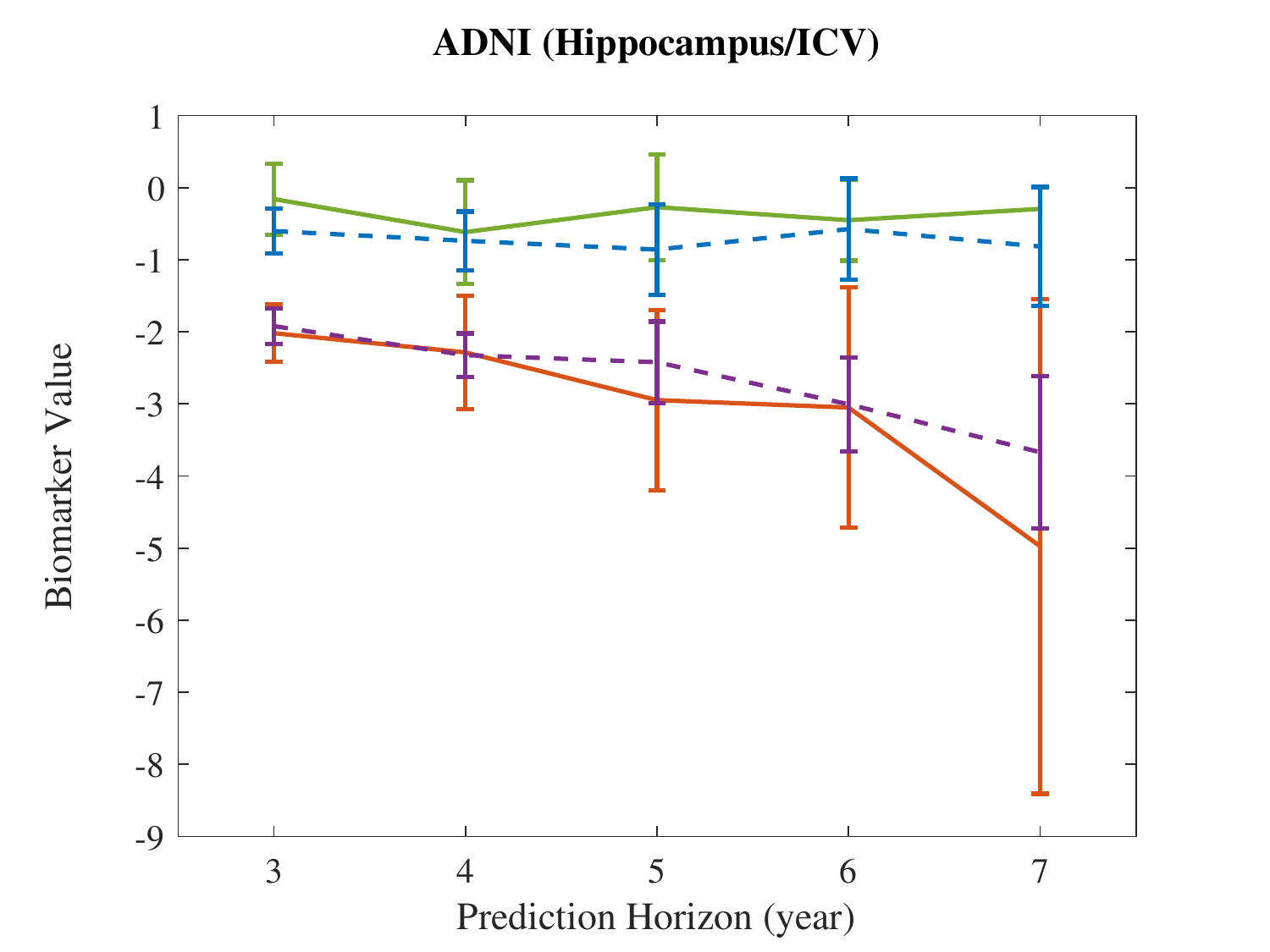}}
\end{subfigure}
\begin{subfigure}[t]{0.54\textwidth}
\vspace{0.3cm}
\raisebox{-\height}{\includegraphics[scale=0.6]{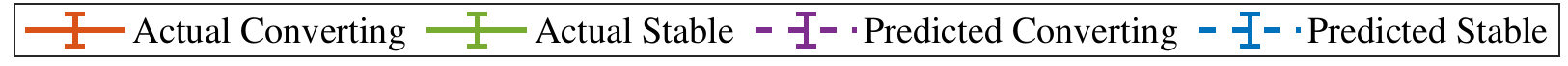}}
\vspace{0.1cm}
\end{subfigure}
\caption{Prediction results for the ADNI test subjects per visit using the trained CAR-GRU models. The error bars are calculated based on a 95\% confidence interval for population standard deviation per visit.}
\label{fig_adni_traject}
\end{figure*}

\begin{figure*}[t]
\centering
\begin{subfigure}[t]{0.505\textwidth}
\raisebox{-\height}{\includegraphics[scale=0.6]{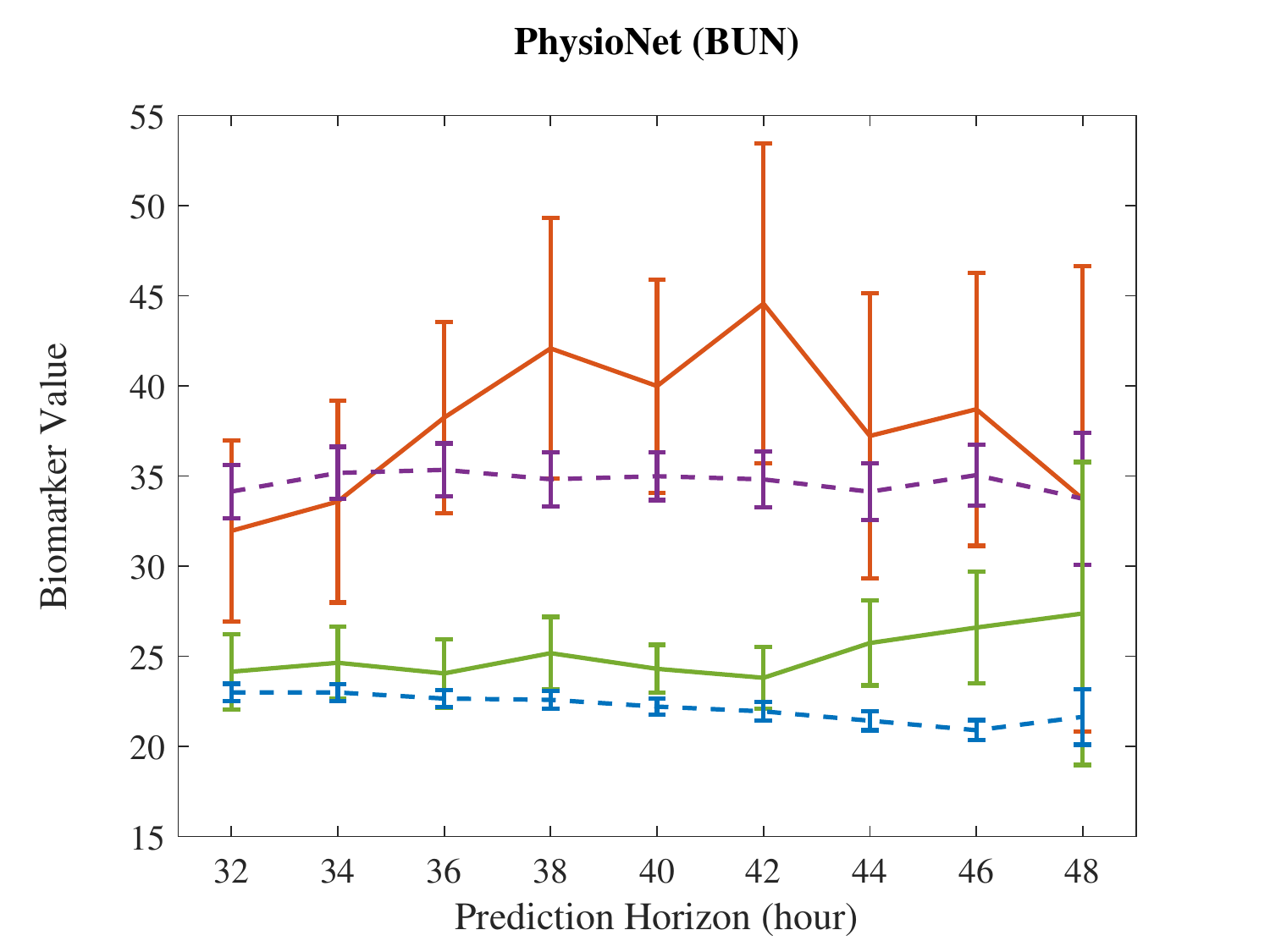}}
\end{subfigure}
\begin{subfigure}[t]{0.485\textwidth}
\raisebox{-\height}{\includegraphics[scale=0.6]{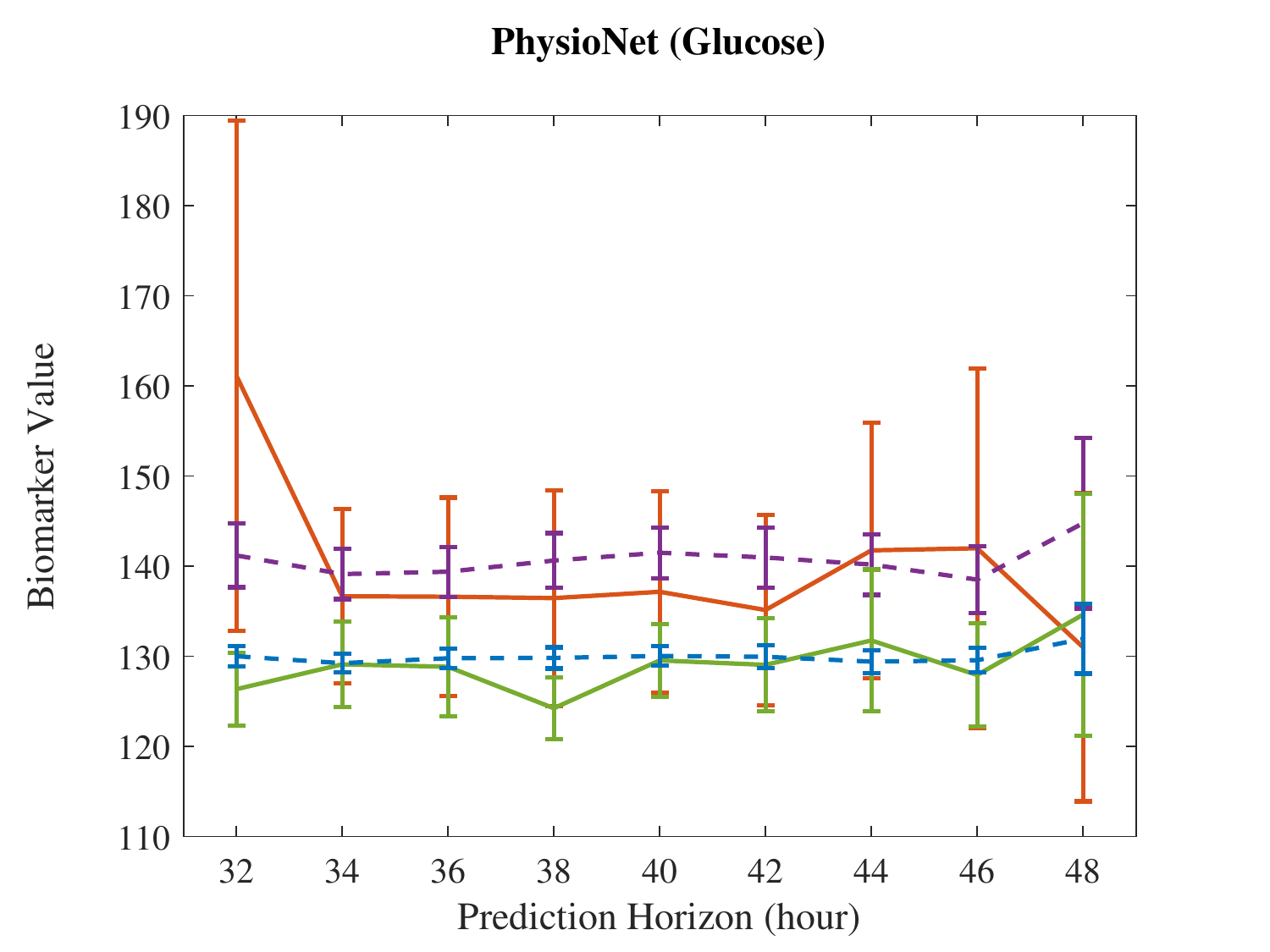}}
\end{subfigure}
\begin{subfigure}[t]{0.505\textwidth}
\raisebox{-\height}{\includegraphics[scale=0.6]{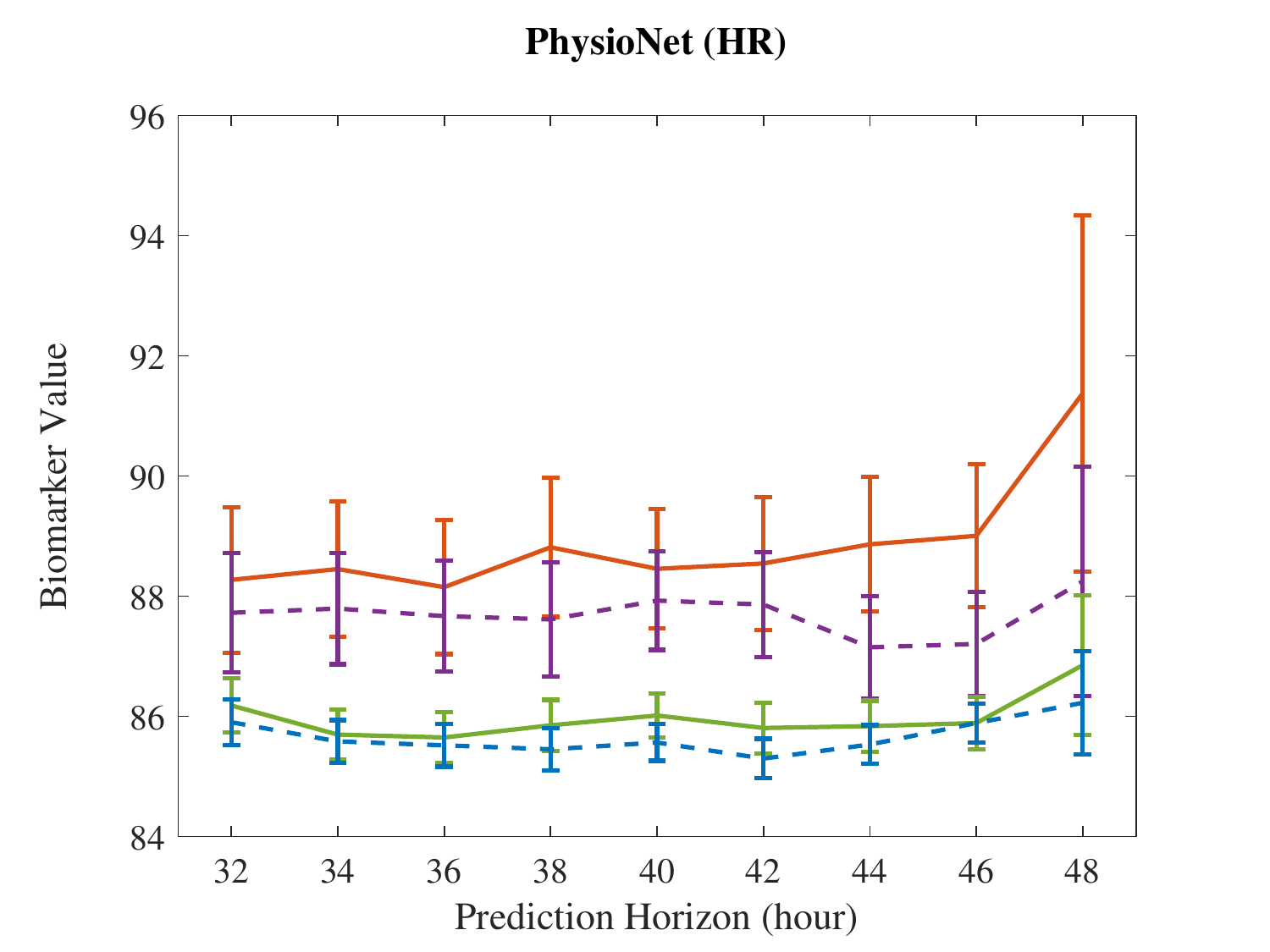}}
\end{subfigure}
\begin{subfigure}[t]{0.485\textwidth}
\raisebox{-\height}{\includegraphics[scale=0.6]{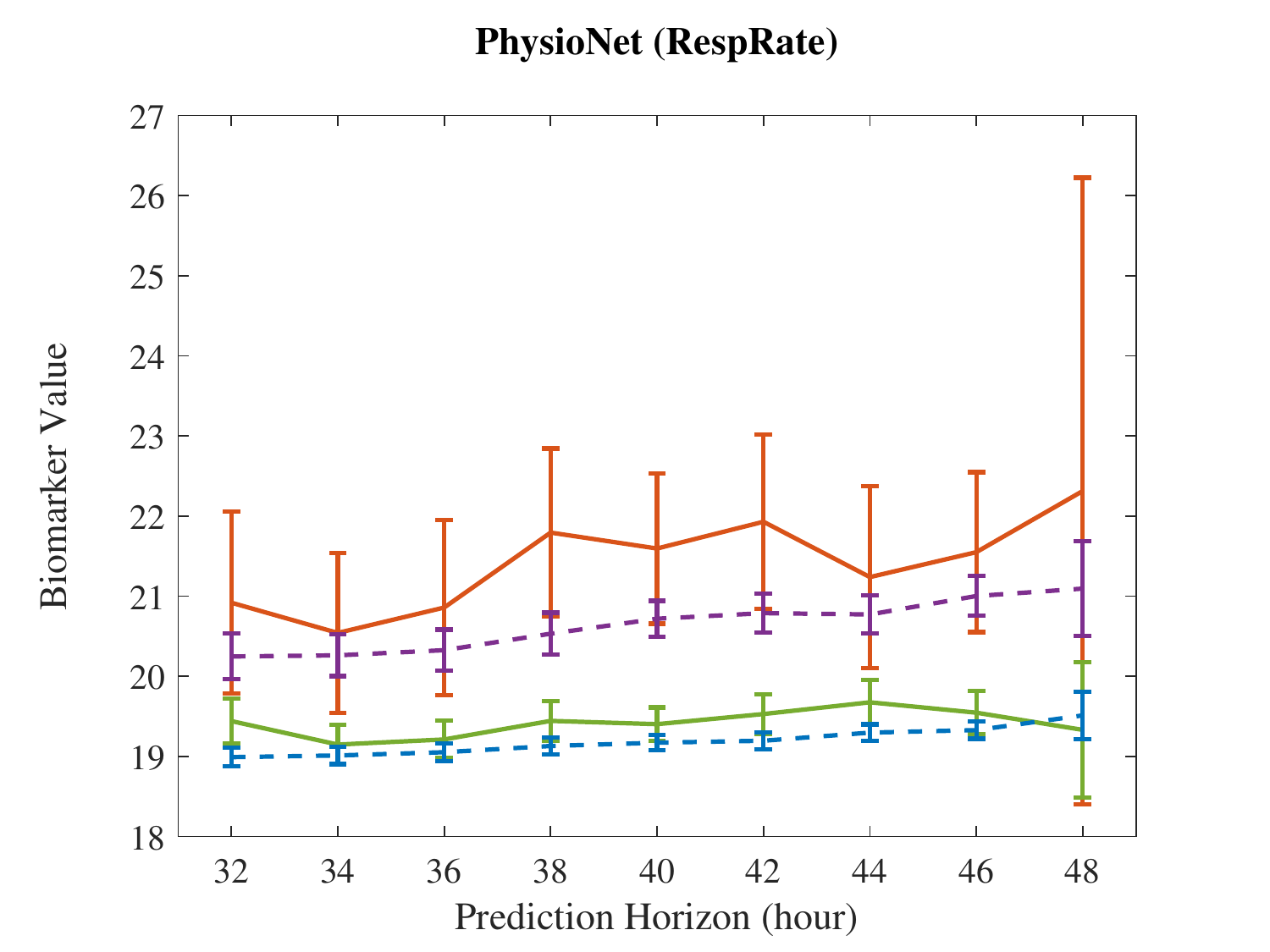}}
\end{subfigure}
\begin{subfigure}[t]{0.54\textwidth}
\vspace{0.3cm}
\raisebox{-\height}{\includegraphics[scale=0.6]{fig_legend_visit.pdf}}
\vspace{0.1cm}
\end{subfigure}
\caption{Prediction results for the PhysioNet test subjects per visit using the trained CAR-GRU models. The error bars are calculated based on a 95\% confidence interval for population standard deviation per visit.}
\label{fig_physio_traject}
\end{figure*}

As the last experiment, we investigate the feature trajectory prediction and discrimination capabilities of the optimal model. To do so, the trained CAR-GRU models are applied to the first 2 years of data from ADNI subjects and to the first 30 hours of data from PhysioNet patients to predict later follow-ups of the measurements for both stable and converting cases. Figures \ref{fig_adni_traject} and \ref{fig_physio_traject} display the predicted trajectories of four important biomarkers from each of the test datasets using the trained CAR-GRU models. As can be seen, the prediction errors stay low in the defined horizon for both stable and converting groups. Moreover, some biomarkers such as the mini-mental state exam (MMSE) cognitive score, the normalized hippocampal volume of the T1-weighted MRI scan (Hippocampus/ICV), blood urea nitrogen (BUN), and heart rate (HR) become more abnormal in the disease course of the converting cases, which can help us to better distinguish between the two groups.

\subsection{General Discussion and Conclusion}

In this paper, a novel deep learning method was proposed for modeling multiple temporal features in sporadic data using an integration of a continuous-time autoregressive model and an RNN to handle irregularity and asynchronicity of measurements and capture long-term dependencies. The model was applied to multivariate time-series regression within two sporadic medical datasets and the obtained results showed that the CAR-GRU method achieved a lower generalization error in predicting the feature values compared to the alternatives.

One of the benefits of the proposed model is the generalizability of the architecture where any type of discrete-time models such as RNNs, and continuous-time models such as Gaussian processes \cite{Futoma2017} can be utilized. Nevertheless, a CAR(1) model with a linear transformation allows for its simulation as a simple neural network modulated by time lags, and hence, its application in deep learning frameworks and architectures including convolutional layers.

The flexibility of the proposed model in generalization to different deep learning architectures provided us an opportunity to compare various types of RNNs applied to the utilized data. The GRU model obtained very decent results while its architecture was simpler than the LSTM one. Moreover, in all experiments, RNN resulted in a larger prediction error compared to LSTM and GRU. There could be two main reasons for such behavior. First, RNNs can suffer from the exploding and vanishing gradients problem as they lack a gating architecture, and therefore, cannot capture long-term dependencies. Second, they use a relatively simple architecture with fewer parameters which can cause underfitting.

We developed an analytical model with a generic solution to deal with irregularity and missing values in temporal data, which explains why the existing deep learning methods \cite{Neil2016,Baytas2017,Che2018,Santeramo2018,Gao2019,Sahin2018,Wu2018} apply exponential or linear time gates to handle the issue as $\bm{h}_{k} = f(\bm{x}_{k}, \bm{h}_{k-1}, g(\Delta t_{k}))$, where $g(\Delta t_{k})$ is the time interval function and can be defined as $[\bm{\Phi}_{h}, \bm{\varsigma}_{h}] \Delta t_{k}$ based on the proposed method. This method can also be compared with ODE-RNNs \cite{Chen2018,Rubanova2019} that attempt to model continuous time-series using RNNs as $\bm{h}_{k} = \bm{h}_{k-1} + f(\bm{h}_{k-1}, t_{k}) \Delta t_{k}$, where $f(\bm{h}(t), t) = d \bm{h}(t) / dt$ is the ODE function and can be approximated by the proposed linearized solution of $[\bm{\Phi}_{h} \bm{h}(t) + \bm{\varsigma}_{h}] \Delta t_{k}$. More interestingly, the proposed way of missing data imputation can be seen as a learning-based alternative to linear interpolation defined by $x(t_{k}) = x(t_{j}) + (t_{k} - t_{j}) dx(t_{j}) / dt$ \cite{Kokic2001}, where the slope at $t_{j}$, i.e., $dx(t_{j}) / dt$, is evaluated by $\varphi x(t_{j}) + \zeta$ in the proposed method. On the other hand, CARRNN can be seen as a deep learning-based, recursive alternative to structural equation modeling (SEM) \cite{Voelkle2012,Van2018} defined by $\bm{h} = \bm{B} \bm{h} + \bm{\psi}$, where $\bm{\psi}$ contains prediction errors and matrix $\bm{B}$ is a concatenation of a diagonal matrix with $\bm{I} + \bm{\Phi} \Delta t_{k}$ elements and a vector containing $\bm{\varsigma} \Delta t_{k}$ values.

The proposed network can be implemented using any deep learning frameworks in three different ways, i.e., a single model with built-in architecture as was proposed in this paper shown in Figure \ref{fig_carrnn}, two separate models with a CAR(1) model proceeding the RNN model or a CAR(1) model preceding the RNN model. Although the separate models provide an opportunity for implementing the methods without changing the RNN architectures, we found no significant improvements in the performance in either case.

\appendices

\section{Backpropagation Through Time} \label{append_gradients}

Let $\mathcal{L} \in \mathbb{R}$ be the loss function defined based on the actual target $\bm{S} \in \mathbb{R}^{Q \times K}$ and the network output $\bm{Y} \in \mathbb{R}^{Q \times K}$, where $Q$ and $K$ stand for the number of output feature nodes and sequence length, respectively. The goal is to derive the partial derivatives of the loss function with respect to the learnable parameters ($\delta \theta = \partial \mathcal{L} / \partial \theta$) using the chain rule. By assuming an L2-norm loss function for regression, the output layer gradients can be obtained as
\begin{gather*}
\textstyle \mathcal{L} = \frac{1}{KQ} \big\Vert \bm{Y} - \bm{S} \big\Vert^{2}_{2} \,, \\
\textstyle \delta \bm{y}_{k} = \frac{1}{2KQ} \big(\bm{y}_{k} - \bm{s}_{k}\big) \,, \\
\delta \bar{\bm{y}}_{k} = \delta \bm{y}_{k} \odot \sigma'_y(\bar{\bm{y}}_{k}) \,,
\end{gather*}
\noindent where $\sigma'(\cdot)$ is the derivative of the activation function $\sigma(\cdot)$, and $\bar{\bm{y}}_{k} \in \mathbb{R}^{Q \times 1}$ is the output layer vector before activation at time point $t_k$.

\subsection{CAR-RNN}

The backpropagation calculations through time for the CARRNN model using an RNN with full gradients are as follows
\begin{gather*}
\delta \bm{h}_{k} = \bm{W}^{\mathsf{T}}_{y} \delta \bar{\bm{y}}_{k} + \bm{U}^{\mathsf{T}}_{h} \delta \tilde{\bar{\bm{h}}}_{k+1} \,, \\
\delta \tilde{\bm{h}}_{k} = \big[\bm{I}_M + (\Delta t_{k} - \tau) \bm{\Phi}^{\mathsf{T}}_{h}\big] \delta \bm{h}_{k} \,, \\
\delta \tilde{\bar{\bm{h}}}_{k} = \delta \tilde{\bm{h}}_{k} \odot \sigma'_{h}(\tilde{\bar{\bm{h}}}_{k}) \,, \\
\delta \bm{x}_{k} = \bm{W}^{\mathsf{T}}_{h} \delta \tilde{\bar{\bm{h}}}_{k} \,,
\end{gather*}
\noindent where $\tilde{\bar{\bm{h}}}_{k} \in \mathbb{R}^{M \times 1}$ is the regularized recurrent vector before activation about time point $t_k$. Finally, the gradients of the loss function with respect to the learnable parameters are obtained as
\begin{gather*}
\textstyle \delta \bm{W}_{y} = \sum_{k} \delta \bar{\bm{y}}_{k} \bm{h}^{\mathsf{T}}_{k} \,, \\
\textstyle \delta \bm{b}_{y} = \sum_{k} \delta \bar{\bm{y}}_{k} \,, \\
\textstyle \delta \bm{W}_{h} = \sum_{k} \delta \tilde{\bar{\bm{h}}}_{k} \bm{x}^{\mathsf{T}}_{k} \,, \\
\textstyle \delta \bm{U}_{h} = \sum_{k} \delta \tilde{\bar{\bm{h}}}_{k+1} \bm{h}^{\mathsf{T}}_{k} \,, \\
\textstyle \delta \bm{b}_{h} = \sum_{k} \delta \tilde{\bar{\bm{h}}}_{k} \,, \\
\textstyle \delta \bm{\Phi}_{h} = \sum_{k} (\Delta t_{k} - \tau) \delta \bm{h}_{k} \tilde{\bm{h}}^{\mathsf{T}}_{k} \,, \\
\textstyle \delta \bm{\varsigma}_{h} = \sum_{k} (\Delta t_{k} - \tau) \delta \bm{h}_{k} \,.
\end{gather*}

Note that the abovementioned gradients are obtained assuming that the utilized data is complete. Therefore, they need to be modified properly according to the explanations in Section \ref{sec_fill} in the cases when the data after binning includes any missing values.

\subsection{CAR-LSTM}
The backpropagation calculations through time for the CARRNN model using a peephole LSTM unit with full gradients are as follows
\begin{gather*}
\resizebox{0.98\hsize}{!}{$\delta \bm{h}_{k} = \bm{W}^{\mathsf{T}}_{y} \delta \bar{\bm{y}}_{k} + \bm{U}^{\mathsf{T}}_{f} \delta \tilde{\bar{\bm{f}}}_{k+1} + \bm{U}^{\mathsf{T}}_{i} \delta \tilde{\bar{\bm{i}}}_{k+1} + \bm{U}^{\mathsf{T}}_{z} \delta \tilde{\bar{\bm{z}}}_{k+1} + \bm{U}^{\mathsf{T}}_{o} \delta \tilde{\bar{\bm{o}}}_{k+1}$} \,, \\
\delta \tilde{\bm{h}}_{k} = \big[\bm{I}_M + (\Delta t_{k} - \tau) \bm{\Phi}^{\mathsf{T}}_{h}\big] \delta \bm{h}_{k} \,, \\
\delta \tilde{\bm{o}}_{k} = \delta \tilde{\bm{h}}_{k} \odot \tilde{\bm{c}}_{k} \,, \\
\delta \tilde{\bar{\bm{o}}}_{k} = \delta \tilde{\bm{o}}_{k} \odot \sigma'_{g}(\tilde{\bar{\bm{o}}}_{k}) \,, \\
\delta \tilde{\bm{c}}_{k} = \delta \tilde{\bm{h}}_{k} \odot \tilde{\bm{o}}_{k} \,, \\
\delta \bm{c}_{k} = \bm{V}^{\mathsf{T}}_{f} \delta \tilde{\bar{\bm{f}}}_{k+1} + \bm{V}^{\mathsf{T}}_{i} \delta \tilde{\bar{\bm{i}}}_{k+1} + \bm{V}^{\mathsf{T}}_{o} \delta \tilde{\bar{\bm{o}}}_{k} + \delta \tilde{\bar{\bm{c}}}_{k+1} \odot \tilde{\bm{f}}_{k+1} \,, \\
\delta \tilde{\bar{\bm{c}}}_{k} = \big[\bm{I}_M + (\Delta t_{k} - \tau) \bm{\Phi}^{\mathsf{T}}_{c}\big] \delta \bm{c}_{k} + \delta \tilde{\bm{c}}_{k} \odot \sigma'_{h}(\tilde{\bar{\bm{c}}}_{k}) \,, \\
\delta \tilde{\bm{z}}_{k} = \delta \tilde{\bar{\bm{c}}}_{k} \odot \tilde{\bm{i}}_{k} \,, \\
\delta \tilde{\bar{\bm{z}}}_{k} = \delta \tilde{\bm{z}}_{k} \odot \sigma'_{c}(\tilde{\bar{\bm{z}}}_{k}) \,, \\
\delta \tilde{\bm{i}}_{k} = \delta \tilde{\bar{\bm{c}}}_{k} \odot \tilde{\bm{z}}_{k} \,, \\
\delta \tilde{\bar{\bm{i}}}_{k} = \delta \tilde{\bm{i}}_{k} \odot \sigma'_{g}(\tilde{\bar{\bm{i}}}_{k}) \,, \\
\delta \tilde{\bm{f}}_{k} = \delta \tilde{\bar{\bm{c}}}_{k} \odot \bm{c}_{k-1} \,, \\
\delta \tilde{\bar{\bm{f}}}_{k} = \delta \tilde{\bm{f}}_{k} \odot \sigma'_{g}(\tilde{\bar{\bm{f}}}_{k}) \,, \\
\delta \bm{x}_{k} = \bm{W}^{\mathsf{T}}_{f} \delta \tilde{\bar{\bm{f}}}_{k} + \bm{W}^{\mathsf{T}}_{i} \delta \tilde{\bar{\bm{i}}}_{k} + \bm{W}^{\mathsf{T}}_{z} \delta \tilde{\bar{\bm{z}}}_{k} + \bm{W}^{\mathsf{T}}_{o} \delta \tilde{\bar{\bm{o}}}_{k} \,,
\end{gather*}
\noindent where $\{\tilde{\bar{\bm{f}}}_{k},\tilde{\bar{\bm{i}}}_{k},\tilde{\bar{\bm{z}}}_{k},\tilde{\bar{\bm{c}}}_{k},\tilde{\bar{\bm{o}}}_{k}\} \in \mathbb{R}^{M \times 1}$ are the regularized vectors of forget gate, input gate, modulation gate, cell state, and output gate before activation about time point $t_{k}$, respectively. Finally, the gradients of the loss function with respect to the learnable parameters are obtained as
\begin{gather*}
\textstyle \delta \bm{W}_{\pi \in \{f,i,z,o\}} = \sum_{k} \delta \tilde{\bar{\bm{\pi}}}_{k} \bm{x}^{\mathsf{T}}_{k} \,, \\
\textstyle \delta \bm{U}_{\pi \in \{f,i,z,o\}} = \sum_{k} \delta \tilde{\bar{\bm{\pi}}}_{k+1} \bm{h}^{\mathsf{T}}_{k} \,, \\
\textstyle \delta \bm{V}_{\pi \in \{f,i\}} = \sum_{k} \mathrm{diag}(\delta \tilde{\bar{\bm{\pi}}}_{k+1} \odot \bm{c}_{k}) \,, \\
\textstyle \delta \bm{V}_{o} = \sum_{k} \mathrm{diag}(\delta \tilde{\bar{\bm{o}}}_{k} \odot \bm{c}_{k}) \,, \\
\textstyle \delta \bm{b}_{\pi \in \{f,i,z,o\}} = \sum_{k} \delta \tilde{\bar{\bm{\pi}}}_{k} \,, \\
\textstyle \delta \bm{\Phi}_{\pi \in \{h,c\}} = \sum_{k} (\Delta t_{k} - \tau) \delta \bm{\pi}_{k} \tilde{\bar{\bm{\pi}}}^{\mathsf{T}}_{k} \,, \\
\textstyle \delta \bm{\varsigma}_{\pi \in \{h,c\}} = \sum_{k} (\Delta t_{k} - \tau) \delta \bm{\pi}_{k} \,.
\end{gather*}

\subsection{CAR-GRU}
The backpropagation calculations through time for the CARRNN model using a GRU with full gradients are as follows
\begin{gather*}
\delta \bm{h}_{k} = \bm{W}^{\mathsf{T}}_{y} \delta \bar{\bm{y}}_{k} + \bm{U}^{\mathsf{T}}_{z} \delta \tilde{\bar{\bm{z}}}_{k+1} + \bm{U}^{\mathsf{T}}_{r} \delta \tilde{\bar{\bm{r}}}_{k+1} \\
+ \delta \tilde{\bm{h}}_{k+1} \odot \tilde{\bm{z}}_{k+1} + \tilde{\bm{r}}_{k+1} \odot (\bm{U}^{\mathsf{T}}_{c} \delta \tilde{\bar{\bm{c}}}_{k+1}) \,, \\
\textstyle \delta \tilde{\bm{h}}_{k} = \big[\bm{I}_M + (\Delta t_{k} - \tau) \bm{\Phi}^{\mathsf{T}}_{h}\big] \delta \bm{h}_{k} \,, \\
\delta \tilde{\bm{c}}_{k} = \delta \tilde{\bm{h}}_{k} \odot (1 - \tilde{\bm{z}}_{k}) \,, \\
\delta \tilde{\bar{\bm{c}}}_{k} = \delta \tilde{\bm{c}}_{k} \odot \sigma'_{h}(\tilde{\bar{\bm{c}}}_{k}) \,, \\
\delta \tilde{\bm{r}}_{k} = \bm{h}_{k-1} \odot (\bm{U}^{\mathsf{T}}_{c} \delta \tilde{\bar{\bm{c}}}_{k}) \,, \\
\delta \tilde{\bar{\bm{r}}}_{k} = \delta \tilde{\bm{r}}_{k} \odot \sigma'_{g}(\tilde{\bar{\bm{r}}}_{k}) \,, \\
\delta \tilde{\bm{z}}_{k} = \delta \tilde{\bm{h}}_{k} \odot (\bm{h}_{k-1} - \tilde{\bm{c}}_{k}) \,, \\
\delta \tilde{\bar{\bm{z}}}_{k} = \delta \tilde{\bm{z}}_{k} \odot \sigma'_{g}(\tilde{\bar{\bm{z}}}_{k}) \,, \\
\delta \bm{x}_{k} = \bm{W}^{\mathsf{T}}_{z} \delta \tilde{\bar{\bm{z}}}_{k} + \bm{W}^{\mathsf{T}}_{r} \delta \tilde{\bar{\bm{r}}}_{k} + \bm{W}^{\mathsf{T}}_{c} \delta \tilde{\bar{\bm{c}}}_{k} \,,
\end{gather*}
\noindent where $\{\tilde{\bar{\bm{z}}}_{k},\tilde{\bar{\bm{r}}}_{k},\tilde{\bar{\bm{c}}}_{k}\} \in \mathbb{R}^{M \times 1}$ are the regularized vectors of update gate, reset gate, and candidate state before activation about time point $t_{k}$, respectively. Finally, the gradients of the loss function with respect to the learnable parameters are obtained as
\begin{gather*}
\textstyle \delta \bm{W}_{\pi \in \{z,r,c\}} = \sum_{k} \delta \tilde{\bar{\bm{\pi}}}_{k} \bm{x}^{\mathsf{T}}_{k} \,, \\
\textstyle \delta \bm{U}_{\pi \in \{z,r\}} = \sum_{k} \delta \tilde{\bar{\bm{\pi}}}_{k+1} \bm{h}^{\mathsf{T}}_{k} \,, \\
\textstyle \delta \bm{U}_{c} = \sum_{k} \delta \tilde{\bar{\bm{c}}}_{k+1} (\tilde{\bm{r}}_{k+1} \odot \bm{h}_{k})^{\mathsf{T}} \,, \\
\textstyle \delta \bm{b}_{\pi \in \{z,r,c\}} = \sum_{k} \delta \tilde{\bar{\bm{\pi}}}_{k} \,, \\
\textstyle \delta \bm{\Phi}_{h} = \sum_{k} (\Delta t_{k} - \tau) \delta \bm{h}_{k} \tilde{\bm{h}}^{\mathsf{T}}_{k} \,, \\
\textstyle \delta \bm{\varsigma}_{h} = \sum_{k} (\Delta t_{k} - \tau) \delta \bm{h}_{k} \,.
\end{gather*}

\section*{Acknowledgment}

This project has received funding from the European Union's Horizon 2020 research and innovation programme under the Marie Sk{\l}odowska-Curie grant agreement No. 721820, No. 643417, No. 681043 and No. 825664, and VELUX FONDEN and Innovation Fund Denmark under the grant number 9084-00018B.

Data collection and sharing for this project was funded by the Alzheimer's Disease Neuroimaging Initiative (ADNI) (National Institutes of Health Grant U01 AG024904) and DOD ADNI (Department of Defense award number W81XWH-12-2-0012). ADNI is funded by the National Institute on Aging, the National Institute of Biomedical Imaging and Bioengineering, and through generous contributions from the following: AbbVie, Alzheimer's Association; Alzheimer's Drug Discovery Foundation; Araclon Biotech; BioClinica, Inc.; Biogen; Bristol-Myers Squibb Company; CereSpir, Inc.; Cogstate; Eisai Inc.; Elan Pharmaceuticals, Inc.; Eli Lilly and Company; EuroImmun; F. Hoffmann-La Roche Ltd. and its affiliated company Genentech, Inc.; Fujirebio; GE Healthcare; IXICO Ltd.; Janssen Alzheimer Immunotherapy Research \& Development, LLC.; Johnson \& Johnson Pharmaceutical Research \& Development LLC.; Lumosity; Lundbeck; Merck \& Co., Inc.; Meso Scale Diagnostics, LLC.; NeuroRx Research; Neurotrack Technologies; Novartis Pharmaceuticals Corporation; Pfizer Inc.; Piramal Imaging; Servier; Takeda Pharmaceutical Company; and Transition Therapeutics. The Canadian Institutes of Health Research is providing funds to support ADNI clinical sites in Canada. Private sector contributions are facilitated by the Foundation for the National Institutes of Health (www.fnih.org). The grantee organization is the Northern California Institute for Research and Education, and the study is coordinated by the Alzheimer's Therapeutic Research Institute at the University of Southern California. ADNI data are disseminated by the Laboratory for Neuro Imaging at the University of Southern California.

\bibliography{references}
\bibliographystyle{IEEEtran}

\end{document}